\documentclass[10pt,journal,compsoc]{IEEEtran}
\ifCLASSOPTIONcompsoc
  \usepackage[nocompress]{cite}
\else
  \usepackage{cite}
\fi
\ifCLASSINFOpdf
\usepackage[pdftex]{graphicx}
\else
\fi
\usepackage{amssymb}
\usepackage{algorithm}
\usepackage{algorithmicx}
\usepackage{algpseudocode}
\usepackage{subfig}
\usepackage{url}

\def\eg{\emph{e.g}., } 
\def\ie{\emph{i.e}., } 
\def\etal{\emph{et al}.}

\usepackage{multirow}
\usepackage{booktabs}
\usepackage{array}
\newcolumntype{P}[1]{>{\centering\arraybackslash}p{#1}}

\usepackage[pagebackref,breaklinks,colorlinks]{hyperref}
\usepackage[normalem]{ulem} 
\usepackage{cuted}
\usepackage[capitalize]{cleveref}
\crefname{section}{Sec.}{Secs.}
\crefname{section}{Section}{Sections}
\crefname{table}{Table}{Tables}
\crefname{table}{Tab.}{Tabs.}
\crefname{lstlisting}{List.}{Lists.}

\usepackage{listings}

\usepackage{xcolor}
\usepackage{bm}
\usepackage{todonotes}

\definecolor{codegreen}{rgb}{0,0.6,0}
\definecolor{codegray}{rgb}{0.5,0.5,0.5}
\definecolor{codepurple}{rgb}{0.58,0,0.82}
\definecolor{backcolour}{rgb}{0.95,0.95,0.92}

\usepackage{etoolbox}
\makeatletter
\AfterEndEnvironment{algorithm}{\let\@algcomment\relax}
\AtEndEnvironment{algorithm}{\kern2pt\hrule\relax\vskip3pt\@algcomment}
\let\@algcomment\relax
\newcommand\algcomment[1]{\def\@algcomment{\footnotesize#1}}
\renewcommand\fs@ruled{\def\@fs@cfont{\bfseries}\let\@fs@capt\floatc@ruled
  \def\@fs@pre{\hrule height.8pt depth0pt \kern2pt}%
  \def\@fs@post{}%
  \def\@fs@mid{\kern2pt\hrule\kern2pt}%
  \let\@fs@iftopcapt\iftrue}
\makeatother

\begin{document}
%
\title{Virtual Category Learning: A Semi-Supervised Learning Method for Dense Prediction with Extremely Limited Labels}
%
%
%
%

\author{Changrui~Chen,
        Jungong~Han*,
        Kurt~Debattista
\IEEEcompsocitemizethanks{\IEEEcompsocthanksitem C. Chen (changrui.chen@warwick.ac.uk) and K. Debattista are with WMG, University of Warwick, UK.\protect
\IEEEcompsocthanksitem J. Han, the corresponding author, is with University of Sheffield and University of Warwick, UK.}
}

%
%

\markboth{Journal of \LaTeX\ Class Files,~Vol.~14, No.~8, August~2015}%
{Shell \MakeLowercase{\textit{et al.}}: Bare Demo of IEEEtran.cls for Computer Society Journals}
%



\IEEEtitleabstractindextext{%
\begin{abstract}
Due to the costliness of labelled data in real-world applications, semi-supervised learning, underpinned by pseudo labelling, is an appealing solution. However, handling confusing samples is nontrivial: discarding valuable confusing samples would compromise the model generalisation while using them for training would exacerbate the issue of confirmation bias caused by the resulting inevitable mislabelling. To solve this problem, this paper proposes to use confusing samples proactively without label correction. Specifically, a Virtual Category (VC) is assigned to each confusing sample in such a way that it can safely contribute to the model optimisation even without a concrete label. This provides an upper bound for inter-class information sharing capacity, which eventually leads to a better embedding space. Extensive experiments on two mainstream dense prediction tasks --- semantic segmentation and object detection, demonstrate that the proposed VC learning significantly surpasses the state-of-the-art, especially when only very few labels are available. Our intriguing
findings highlight the usage of VC learning in dense vision tasks.
\end{abstract}

}

\maketitle

\IEEEdisplaynontitleabstractindextext

%
\IEEEpeerreviewmaketitle

\IEEEraisesectionheading{\section{Introduction}\label{sec:intro}}

\

%
%
%
%

\IEEEPARstart{D}{eep} Learning solutions are significantly disadvantaged by the expensive labelling cost of large-scale datasets, especially on dense prediction tasks. Annotating the data for tasks such as object detection and semantic segmentation takes substantially longer when compared to non-dense applications such as image classification. Though crowd-sourcing platforms facilitate  data labelling for many common application scenarios, such as autonomous driving, data labelling in scientific applications usually requires expert labelling, which is not readily available. Semi-supervised learning, which makes use of limited labelled data in combination with large amounts of unlabelled data for training, has shown great potential to reduce the reliance on large amounts of data labelling~\cite{liu2021unbiased,berthelot2019mixmatch,sohn2020fixmatch,li2021comatch}. Diminishing the performance gap between fully- and semi-supervised methods enables the introduction of deep-learning models to more application topics. In existing semi-supervised learning frameworks, a challenging problem is: \textit{how to best utilise the unlabelled data.}

Pseudo labelling (PL)~\cite{lee2013pseudolabel} has recently emerged as a solution to the above problem~\cite{liu2021unbiased,sohn2020fixmatch} and achieved state-of-the-art performance. Here, unlabelled data are automatically annotated by the model itself~\cite{sohn2020a} (or via an exponential moving average version~\cite{liu2021unbiased}) and then fed back to re-optimise the model. Despite their preliminary success, existing PL-based semi-supervised methods are not good at or even incapable of dealing with the extremely-scarce label setting. It motivates us to study the limited supervised case, which is imperative in real-life application scenarios.

\begin{figure}[t]
\centering
\subfloat[]{
\includegraphics[width=0.49\columnwidth]{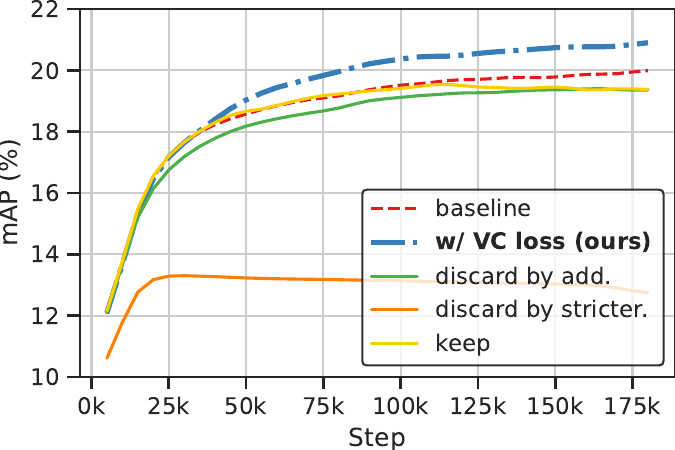}
\label{fig:intro-a}
}
\subfloat[]{
\includegraphics[width=0.47\columnwidth]{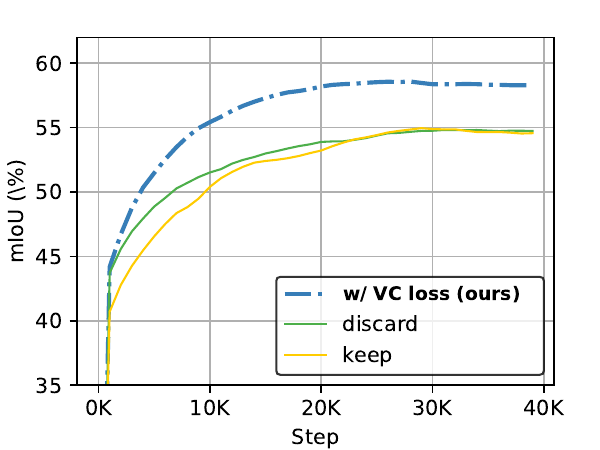}
\label{fig:intro-b}
}
\caption{a): The mAP of a semi-supervised detector~\cite{liu2021unbiased} with a preset confidence score filtering on 1\% labelled MS COCO~\cite{lin2014microsoft}. The mAP sees a decrease with all strategies for dealing with confusing samples (\eg the bear-like dog at the right) except for the VC learning. The additional filtering mechanism (add.) is the temporal stability verification proposed in this paper. Stricter filtering (stricter.) raises the threshold in the score filtering from 0.7 to 0.8. b): The mIoU of a semi-supervised semantic segmentor on 1/128 labelled Pascal VOC.}\label{fig:intro}
\end{figure}

Typically, due to the limited diversity of training samples in a very small available set of labelled data, the non-optimal decision boundary usually leads to an indecisive decision on some unseen confusing samples when inferring their pseudo labels. In PL, two strategies are usually adopted to deal with confusing samples: a) discarding all of them using a strict filtering mechanism~\cite{sohn2020a}, or b) retaining them with all potential labels~\cite{yang2021interactive}. However, neither of these two options is optimal, especially when the labelled training data are very limited. The value of confusing hard samples is clear to see since hard example mining~\cite{shrivastava2016training} has successfully proven its effectiveness in fully-supervised learning. If all confusing samples are rejected by a strict filtering mechanism, their positive contributions will be wasted, while the remaining well-fitted samples only marginally contribute to performance improvements. On the contrary, simply keeping them all is ineffective due to the involvement of too many incorrect pseudo labels. Arbitrarily optimising semi-supervised detectors with these noisy labels results in confirmation bias issue~\cite{arazo2020pseudolabeling}. To demonstrate these points, we show in~\cref{fig:intro-a} the mean Average Precision (mAP) of a semi-supervised object detector with different strategies on 1\% labelled MS COCO~\cite{lin2014microsoft}. Noticeable performance degradation can be observed when either choosing one stricter filtering mechanism (orange line) or adding an additional one (green line) to reject confusing samples. Likewise, simply keeping all confusing samples (yellow line) pseudo labels also ends up with a decreased mAP since the unreliable pseudo labels aggravate the confirmation bias issue and can result in training collapse. A similar phenomenon can be seen in semantic segmentation (\cref{fig:intro-b}) as well.

In light of the above, efforts have been dedicated to exploring how to correct the biased pseudo labels to utilise the confusing samples efficiently. Existing methods~\cite{li2021comatch} initially investigate relatively straightforward tasks such as classification on CIFAR~\cite{krizhevsky2009learning}. However, promising progress has not yet been made for a complex dense prediction task, such as object detection with extremely small amounts of labelled data. Then, a question arises: \textit{what if we do not discard confusing samples but consider their contributions, which may not necessarily need the concrete label information, during the model training?} This paper answers this question by proposing a novel Virtual Category (VC) learning based on the observation that there is an implicit and safe optimisation direction in PL models for confusing data. 

\begin{figure}[t]
\centering

\includegraphics[width=0.7\columnwidth]{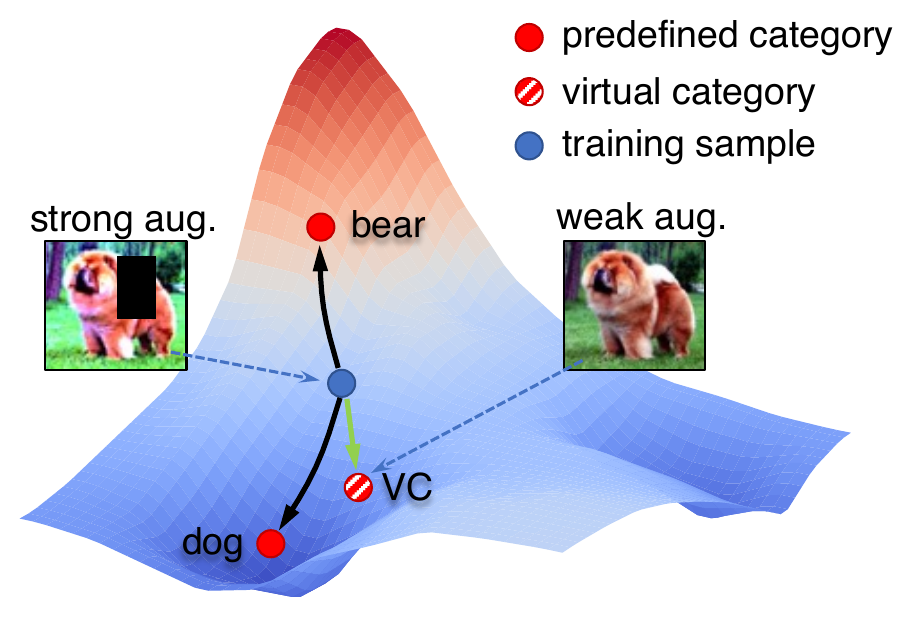}

\caption{Illustration of the basic idea of the virtual category in the manifold of the optimisation space. The peak indicates a high testing error value.}\label{fig:intro-manifold}

\end{figure}

\cref{fig:intro-manifold} provides an example, in which the bear-like dog is a typical confusing sample due to its appearance. For a classification model, the arrow pointing towards the `dog' is the best optimising direction, which results in the smallest testing error value. However, if the doubtful pseudo label is `bear', the incorrect optimising direction would lead to worse performance. We discover that building a Potential Category (PC) set consisting of the possible categories of a confusing sample $x$, compared to determining the exact correct label, is much easier. The remaining task is to find a good optimising direction (labelled VC in \cref{fig:intro-manifold}) for the sample $x$ without the guidance of the categories in the potential category set. Therefore, instead of selecting the correct one from the potential category set, which is usually challenging, we compromise by proposing a VC label to \textit{take the place of all unreliable labels in the potential category set}. A new learning scheme, namely VC learning, allows the model to be optimised with the VC label. By ignoring the categories in the potential category, it will disable the gradient of the corresponding output logits, thus avoiding any wrong optimising direction that would mislead the model. Most importantly, the proposed VC specifies a reasonable upper bound for the inter-class sharing information capcity. Hence, the decision boundary can consistently benefit from the confusing data without suffering from the confirmation bias issue. With regards to the potential category set, we come up with multiple methods to build it. As can be seen in~\cref{fig:intro}, the performance of the model armed with the proposed VC learning (dot-dashed line) sees a significant increase due to the effective use of the confusing samples.

The proposed VC learning is applied to a semi-supervised detector and a semi-supervised segmentor, both with extrmely limited labelled training data, to evaluate its effectiveness and generalisation capability on dense prediction tasks. In the object detection on MS COCO, VC learning achieves 19.46 mAP with only 586 labelled images; this even outperforms some recently published semi-supervised detectors~\cite{sohn2020a,zhou2021instantteaching} with 1000+ labelled images. For semantic segmentation, we developed a powerful and straightforward pseudo-labelling framework. VC learning further boosts the baseline framework to achieve a mIoU of 55.37 on Pascal VOC with only 82 labelled images. It surpasses state-of-the-art methods by a large margin. The contributions of this article are summarised as follows:

\begin{itemize}
\item We take advantage of \textit{confusing samples} with pseudo labels in a semi-supervised manner through VC learning. Our VC learning alleviates the confirmation bias issue caused by confusing samples. It works extremely well when only very limited labelled training data are available. 

\item We theoretically exhibit the feasibility of using VC learning for semi-supervised learning. The findings highlight the need to rethink the usage of confusing samples in semi-supervised tasks.
\item Compared to our previous ECCV oral paper~\cite{chen2022semisupervised}, we extend VC learning to semi-supervised semantic segmentation. More methods for the potential category set creation and an additional loss function form are introduced, \ie the mean squared error for VC learning. On top of it, we also propose a new module to generate the virtual weight in VC learning. The experiments show that VC learning can be well applied in semantic segmentation. 
\item We incorporate VC learning into a unified pseudo-labelling framework for semi-supervised learning, which can deal with multiple dense prediction tasks, including semantic segmentation and object detection. The proposed framework surpasses state-of-the-art methods by a significant margin on ALL tasks, which verifies the generalisation of VC learning.

\end{itemize}


\section{Related Works}
\label{sec:related}

In this section, a literature review on the semi-supervised learning and the downstream tasks --- segmentation and detection --- is conducted.

\subsection{Semi-supervised Learning}

In the last few years, numerous deep backbones and modules training under a fully-supervised scheme have been proposed. VGGNet~\cite{simonyan2015very} adopted $3\times3$ convolution layer as the main model component. GoogLeNet~\cite{szegedy2015going} proposed the network-in-network structure for the first time, which allows scaling the width and depth of a convolution neural network (CNN). ResNet~\cite{he2016deep}, proposed the residual block, which made it possible to optimise a very deep neural network. Recently, inspired by the attention module in neural language processing~\cite{vaswani2017attention}, the vision transformer~\cite{dosovitskiy2021an} appealed to many researchers. Fully-supervised training is very close to the ability of human beings in many vision tasks~\cite{he2015delving}. However, in other research and application areas, it is usually difficult to build large-scale labelled datasets such as ImageNet~\cite{deng2009imagenet} and MS COCO~\cite{lin2014microsoft} to satisfy the training of fully supervised models. Semi-supervised learning, such as MeanTeacher~\cite{tarvainen2017mean} and FixMatch~\cite{sohn2020fixmatch} etc., tackles this issue.

Semi-supervised learning is a training scheme that uses only a small amount of labelled data and a large amount of unlabelled data to train a model. It can be grouped into three main paradigms: a) generative models, b) graph-based methods, and c) pseudo-labelling models. Several unsupervised generative models were extended to solve the semi-supervised problems. For example, the stacked semi-supervised generative model proposed by Kingma \etal~\cite{kingma2014semisupervised} appended a generative classifier to the latent representation produced by the encoder to enable variational autoencoder to tackle semi-supervised classification. The feature representation would benefit from the reconstruction proxy task of the auto-encoder with the unlabelled data.  Generative Adversarial Network (GAN) has also been considered as semi-supervised learning methods~\cite{odena2016semisupervised}. By assigning a `fake' class to all generated images, unlabelled images in the dataset can be labelled as `non-fake' to train the classifier~\cite{odena2016semisupervised}. In addition to the generative model, some graph-based methods introduced data relationships into semi-supervised training~\cite{li2021comatch,luo2018smooth}. The intuitive motivation is that adjacent nodes in an embedding graph should have similar representations. Recently proposed semi-supervised methods mainly focus on the teacher-student framework via pseudo labelling~\cite{sohn2020fixmatch,berthelot2019mixmatch}. The overall idea of a teacher-student framework is to let the predictions of the teacher model be the pseudo labels for optimising the student model. It requires models to produce consistent outputs when the inputs are perturbed. Image augmentations, such as flipping, Cutout~\cite{devries2017improved}, or Gaussian Blurring, are usually applied to perturb input images. Some solutions take advantage of adversarial learning and proposed learnable adversarial augmentations~\cite{miyato2019virtual}. The form of teacher model is in a variety of styles. An exponential moving averaged (EMA) version~\cite{tarvainen2017mean} or even the student itself~\cite{sohn2020fixmatch} was investigated to play the role of the teacher. Many different entities of consistency regularisation have been explored. For example, Jeong \etal~\cite{jeong2019consistencybased} tried to minimise the discrepancy between the latent representations of perturbed inputs. Yang \etal~\cite{yang2021interactive} proposed to use the temporal ensembling predictions as the teacher predictions. FixMatch~\cite{sohn2020fixmatch} adopted a weak and a strong augmentation to obtain the predictions of the teacher and student model, respectively. The teacher-student framework, which is used as the baseline in this paper, has proven to be successful in several downstream tasks. Despite their successes, the performances of such systems are far from satisfactory in the real scenario, where an extremely low label ratio, say below 1\%, is provided. 

\subsection{Semantic Segmentation}

Semantic segmentation can be seen as a dense classification task at the pixel level. Most recent segmentation models are inspired by FCN~\cite{long2015fully}. The performance of segmentation models is sensitive to the output resolution. Thus, some following works, such as U-Net~\cite{ronneberger2015unet}, proposed an encoder-decoder framework to increase the output resolution without compromising efficiency. The receptive field is also crucial to segmentation accuracy. Atrous convolution used by the Deeplab series~\cite{chen2015semantic,chen2017rethinking,chen2018deeplab} decently enlarged the receptive field without the aid of stacking large convolutional kernels. The self-attention mechanism enables the segmentor to build long-range connections across the entire images, further exploiting contextual information~\cite{fu2019dual}. Compared to the image-level labels, the cost required by such pixel-level dense labelling makes semi-supervised learning even more crucial. Consistency regularisation is also widely adopted in semi-supervised semantic segmentation. Ouali \etal~\cite{ouali2020semisupervised} proposed to align the output of different decoders or models. The discriminator, which is usually used in adversarial learning, was considered to minimise the distribution distance between the predictions of unlabelled data and the ground truth of labelled data. The teacher-student model has been well introduced to self-supervised segmentation~\cite{zou2021pseudoseg}. However, we discover that its training is very unstable. Thus, this article first investigates and solves this problem to build a strong and stable baseline model and then evaluates the proposed VC learning on it.

\subsection{Object Detection}

\begin{figure*}[t]
  \centering
   \includegraphics[width=0.85\textwidth]{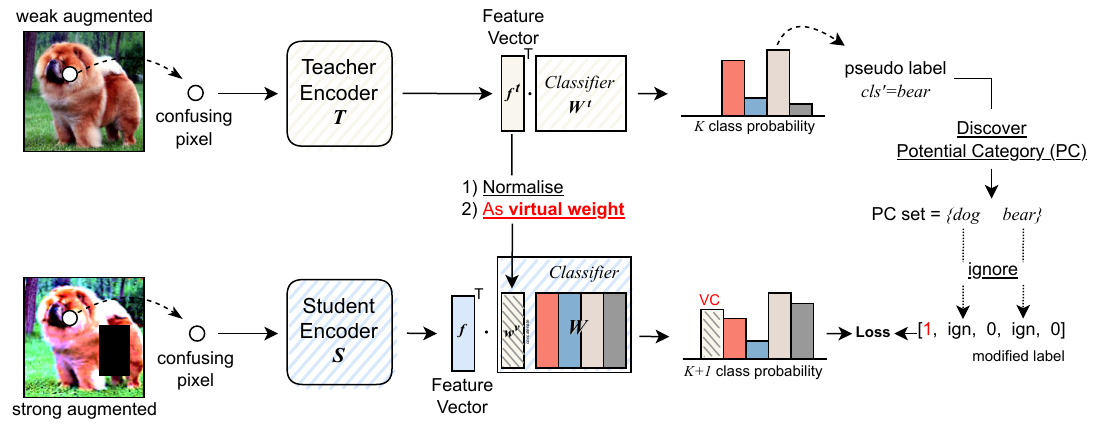}

   \caption{The pipeline of the proposed VC learning when dealing with a confusing sample in semi-supervised one-pixel classification. $T$ is the teacher model. $S$ represents the student model. When training the student classifier with a confusing sample, the weight matrix $W^{s}$ of the student classifier is extended by a virtual weight $w^v$, which is transformed from the corresponding teacher feature vector $\hat f$.}
   \label{fig:framework}

\end{figure*}

 Object Detection, which finds significant applications in downstream tasks, aims to distinguish foreground objects in images or videos and identify them. Object detectors so far can generally be divided into three types: 1) two-stage detectors~\cite{ren2017faster,cai2018cascade}, represented by Faster RCNN~\cite{ren2017faster}; 2) one-stage detectors~\cite{redmon2016you,bochkovskiy2020yolov,lin2020focal,carion2020endtoend}, such as the YOLO series~\cite{redmon2016you,bochkovskiy2020yolov}; and 3) point-based detectors~\cite{duan2019centernet,tian2019fcos,law2018cornernet}, such as Center Net~\cite{duan2019centernet}. The main difference between two-stage and one-stage detectors lies in whether an additional module is used to generate candidate region proposals for classification and localisation. Point-based detectors discard anchor boxes and instead use points and sizes to represent objects. In this paper, Faster RCNN, one of the most widely used detectors, serves as our baseline detector to explore VC learning in semi-supervised object detection(SSOD). SSOD originates from semi-supervised classification, where only a small amount of bounding box labelled data and numerous unlabelled data are available for training a detector. Most of the recently proposed SSOD algorithms followed pseudo-labelling methods. For instance, CSD-SSD~\cite{jeong2019consistencybased} applied consistency regularisation on the predicted classification probability vectors and regression vectors of the input image and its mirror version when dealing with unlabelled images. Several self-supervised detectors~\cite{liu2021unbiased,zhou2021instantteaching,yang2021interactive,sohn2020a}, which provide teacher-produced pseudo-labels for student detectors, have emerged recently.
\\
\\

 Although the teacher-student pseudo labelling technique shows good potential on both tasks, it is still struggling with the confirmation bias issue when training with confusing samples. This paper proactively utilises these confusing samples and alleviates the confirmation bias issue via the novel VC learning method.

\section{Methodology}\label{sec:method}
In this section, the overall problem is first defined. The VC learning and its explanation are subsequently described.

\subsection{Problem Definition} 
In the semi-supervised problem, two data subsets $\mathcal{D}^l$ and $\mathcal{D}^u$ are given for model optimisation, where $\mathcal{D}^l = \{(x^l_n, y^l_n) |_{n=0}^{N^l}\}$ is the subset with ground truth label $y^l$ available, $\mathcal{D}^u = \{x^u_n |_{n=0}^{N^u}\}$ is the unlabelled subset. $N^l$ and $N^u$ are the numbers of labelled and unlabelled data, respectively. This paper mainly investigates the \textbf{limited-supervised learning} problem, \ie, $N^u \gg N^l$, which can be considered as a sort of challenging subproblem of semi-supervised learning. For segmentation, the pixel-level class index $y\in\mathbb{N}_{+}^{H \times W}$  is the label. $y=[a_1, b_1, a_2, b_2, cls]$ is the label of the object detection task, where the first four numbers indicate the coordinates of the top-left and bottom-right points and $cls$ is the index of the category label. This paper follows the teacher-student framework to generate the pseudo label $y'$ of $x^u$ for re-optimising. As shown in~\cref{fig:framework}, two encoders $T$ and $S$, which share the same architecture, are introduced. The parameters of the teacher encoder $T$ are updated by the parameters of the student $S$ with a momentum parameter.

\subsection{Virtual Category Learning}
\label{sec:VCloss}

\begin{figure}[t]
\centering
\includegraphics[width=0.8\columnwidth]{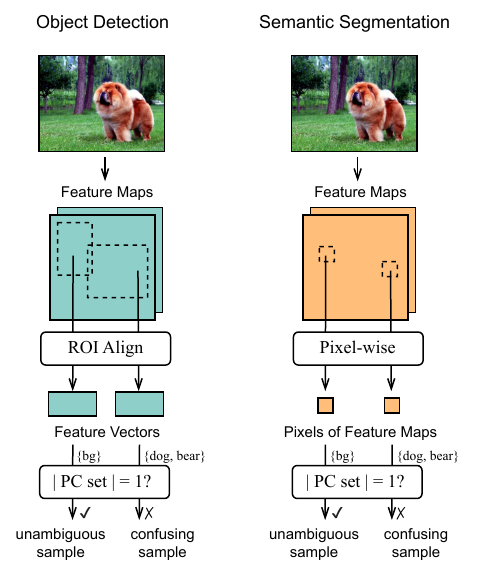}

\caption{Explanation of data sample preparation for different tasks. In object detection, we use ROI Align~\cite{he2017mask} to extract the feature vectors of each region of interest as the data samples. While VC learning is operated in a pixel-wise manner in semantic segmentation.}
\label{fig:data-prepare}

\end{figure}

For ease of understanding, we abstract our VC learning framework for dense vision tasks as a general one-pixel classification task. It can be easily extended to a multi-pixel framework for semantic segmentation and an instance-level framework for object detection as shown in \cref{fig:data-prepare}. In \cref{fig:framework}, two feature encoders $T$ and $S$ first embed the confusing pixel from the weakly and strongly augmented images into the feature space. The linear classifier parameterised by $W^t$ in the teacher branch produces the categorical probability of the input data by performing a matrix multiplication of the feature vector $f^{t\top}$ and the weight matrix $W^t$. The bias parameter is ignored here for simplicity. Typically, the category with the highest probability, \eg $y'=bear$ here, will then be used as the pseudo label for the output of the student branch. However, an incorrect pseudo label may mislead the training.

In this paper, we propose VC learning which modifies the pseudo category label with an additional virtual category to allow the student model to be optimised safely by confusing samples. Once the initial pseudo label $y'=bear$ is obtained, a potential category discovery operation is performed to construct a set $\{dog, bear\}$ for this training sample. We find that the potential category discovery is relatively feasible compared to designing a mapping function $P(y=GT|y'; f)$ to correct wrong pseudo labels, especially when the labelled subset $\mathcal{D}^l$ is much smaller than $\mathcal{D}^u$. The discovery method will be introduced in the following \cref{sec:pcset}.

If $y'=bear$ is the only potential category we can find, this pixel is regarded as an unambiguous sample. If the potential category set contains more than one category, it means that this sample is exactly a confusing sample to the model. In \cref{fig:framework}, a pixel (white circle) of the bear-like dog is an example confusing sample with $\{dog, bear\}$. To encourage the confusing pixel to consistently contribute to the optimisation of the student model $S$ rather than arbitrarily discarding it, the weight matrix $W$ in the student classifier is extended by a `personal' weight vector $w^v$ named \textit{virtual weight}. The ingredient of the virtual weight $w^v$ is the feature vector $f^t$, which is the feature of this confusing sample in the teacher model. Notably, the so-called `personal' virtual weights for different confusing samples are various. With the extended weight matrix, the size of the student classifier output (\ie logits) is therefore increased by 1:
\begin{equation}
	f^\top \cdot [\overbrace{\bm{w^v},w^{0},...,w^{K-1}}^{1+K}]=[\overbrace{\bm{l^v},l^{0},...,l^{K-1}}^{1+K}],
\label{eq:vcfw}
\end{equation} where $K$ is the number of the predefined categories, $l^v$ and $l^{i}$ are the logits of the virtual category and the predefined class $i$ respectively.

\noindent\textbf{Optimisation Objective} To calculate the loss value of the extended logits, the pseudo label is modified by providing a positive label `1' for the virtual category. The training target form is: $[1, 0, ..., ign, ..., ign, 0]$, where `ign' means we ignore that class. Thanks to the virtual category taking on the responsibility of being the target category, the confusing labels in the potential category set can be ignored, thereby avoiding any potential misleading, as it is hard to determine which one is the real ground truth. Thus, the optimisation objective is to get a large logit value for the virtual category and small logit values for the rest of the categories except for those in the potential category set.

\begin{figure}[t]
\centering
\includegraphics[width=0.8\columnwidth]{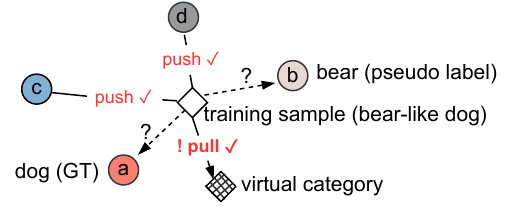}


\caption{Explanation of VC learning in the feature space. The circles are the embeddings of the category cluster centres. The hollow diamond is the embedding of the training sample. The cross-hatch diamond is the embedding of the virtual category.}
\label{fig:explain}

\end{figure}

%

\subsection{Explanation} 

This section describes how the proposed VC learning can be interpreted both from the aspect of the feature space and from the aspect of mathematical feasibility.

\subsubsection{Feature Space}
In the feature space, as shown in~\cref{fig:explain}, let the circles indicate the centres of the predefined categories. Pulling the training sample (diamond) to the circles $a$ or $b$ is risky since we don't know which one is the real ground truth. Given the VC, the decision boundary can still be optimised with VC learning, as it provides a safe optimising direction: pushing the training sample away from the circles $c$ and $d$ and pulling it closer to the diamond of the virtual category. Although one may suspect that our approach looks similar to contrastive learning~\cite{he2020momentum,chen2020a} in terms of the optimisation objective, they differ in several aspects. First, contrastive learning operates before the task-relevant layer (\ie the classifier). As a result, it only drives the backbone encoder to extract better features but does not contribute anything to the task-relevant layer. While our approach acts after the classifier so that the gradient of the virtual category can backpropagate to not only the backbone but also the weight vectors in the classifier. Second, the weight vectors of the other categories in the classifier naturally constitute negative samples such that there is no need to maintain a negative sample pool, which has been a worrying bottleneck for contrastive learning.

\subsubsection{Mathematical Derivation}\label{subsec:math}
To explain our method from the mathematical perspective, we define the loss function of VC learning starting from cross entropy~(CE) loss. Assuming a batch size of 1, the CE loss is:
\begin{equation}
    \label{eq:celoss}
    \mathcal{L}_{CE} = -log(\frac{e^{f^\top \cdot w^{i=GT}}}{\sum_{i=0}^{K} e^{f^\top \cdot w^{i}}}) = log(\sum_{i=0}^{K}e^{l^{i}-l^{GT}}),
\end{equation}
\noindent where $f \in \mathbb{R}^{channel \times 1}$ is the input feature vector of the last linear layer (\ie the classifier), $w^{i} \in \mathbb{R}^{channel\times 1}$ is the corresponding weight vector of the category $i$ in the last linear layer, $l^i=f^\top \cdot  w^i$ is the logit of the category $i$, $K$ is the number of the predefined categories, and $GT$ is the index of the ground truth. 

The intuitive target of minimising CE loss is to get a large logit $l^{GT}$ of the ground truth and small values for the rest of the categories $l^{i \neq GT}$. In a self-supervised scheme, the teacher model infers the pseudo classification label $y'$ of an unlabelled data sample $x^u$. As~\cref{eq:celoss} can be a smooth approximation of the \textit{max} function~\cite{boyd2004convex}, the CE loss with the pseudo label $y'$ can be expressed as:
\begin{equation}
    \label{eq:celoss-2}
    \mathcal{L}_{CE} = log(\sum_{i=0}^{K}e^{l^{i}-l^{y'}}) \approx max_{i \in \{0, ..., K-1\}}(l^i-l^{y'}).
\end{equation} Minimising \cref{eq:celoss-2} is expected to satisfy:
\begin{equation}
    \label{eq:celoss-expect}
(l^{i \neq y'} - l^{y'}) \leq (l^{i=y'} - l^{y'}) \triangleq 0, \hspace{1em} i.e., l^{i \neq y'} \leq l^{y'}.
\end{equation}

\noindent Since $y'$ may not always be correct, when $y'$ is not equal to the real ground truth $GT$, satisfying~\cref{eq:celoss-expect} leads to $l^{GT} \leq l^{y'}$, thereby aggravating the issue typically termed confirmation bias. It can also be explained without the \textit{max} function approximation. The \textit{LogSumExp} is monotonically increasing:
\begin{equation}
	\frac{\partial}{\partial (l^j-l^{y'})} log(\sum_{i=0}^{K}e^{l^{i}-l^{y'}}) = \frac{e^{l^{j}-l^{y'}}}{\sum_{i=0}^{K}e^{l^{i}-l^{y'}}} > 0.
\end{equation}
\noindent Thus, when $y' \neq GT$, the logit $l^{y'}$ will be large and $l^i$ of the other categories (including GT's logit) will be small when $log(\sum_{i=0}^{K}e^{l^{i}-l^{y'}})$ is converged.

With \cref{eq:vcfw}, the loss in VC learning can be defined as follows according to \cref{eq:celoss}:
\begin{equation}
    \label{eq:vcloss}
    \mathcal{L}_{VC-CE} = log(\sum_{i=0,i \notin PC}^{K+1}e^{l^{i}-l^{v}}),
\end{equation} where $PC$ is the potential category set. $i=K$ is the index of the virtual category, \ie $l^{i=K}=l^{v}$. $i \notin PC$ means the labels in the potential category set are ignored in the summation. 

For VC learning, following the derivation of~\cref{eq:celoss-2}, we obtain:
\begin{equation}
    \label{eq:virtualloss-1}
    \mathcal{L}_{VC-CE} = log(\sum_{i=0,i \notin PC}^{K+1}e^{l^{i}-l^{v}}) \approx max_{i \in \{0, ..., K\} \backslash PC}(l^i-l^{v}).
\end{equation} Similar to~\cref{eq:celoss-expect}, minimising~\cref{eq:virtualloss-1} is expecting:
\begin{equation}
    \label{eq:vcloss-expect}
(l^{i \neq v \wedge i \notin PC } - l^{v}) \leq (l^{i=v} - l^{v}) \triangleq 0, \hspace{1em} i.e., l^{i \neq v \wedge i \notin PC} \leq l^{v}.
\end{equation} Comparing~\cref{eq:vcloss-expect} with~\cref{eq:celoss-expect} reveals:

\begin{enumerate}
	\item VC loss first ignores the logits $l^{i \in PC}$ of the confusing labels in the uncertain potential set when satisfying the inequation in \cref{eq:vcloss-expect}, thereby avoiding misleading the training.
	\item Additionally, it provides an alternative upper bound $l^{v}=f^\top \cdot w^{v}$ for all the rest of the logits $l^{i \notin PC}$. The information embedded by the classifier weight vector can be decoupled into two parts: the direction of the vector and the magnitude(norm) of the vector. Since the magnitude is controlled by a norm factor in this paper, VC learning tries to ensure that the cosine similarity of $f$ and $w^v$ is the maximum. The directions of the weight vectors in a linear classifier can represent the information of different categories~\cite{qi2018lowshot}. Thus, $l^{v}=f^\top \cdot w^{v}$ should be larger since $w^v$ is obtained from $f^t$, which is the feature in the teacher of the exact same data sample of feature $f$. $f$ and $f^t$ share a lot of information, thus leading to the largest cosine similarity. Consequently, $l^{v}$ can be a meaningful upper bound for all the rest logits $l^{i \notin PC}$. The shared information between $f$ and $f^t$ is the upper bound of intra-class information sharing capacity.
\end{enumerate}

The VC learning is applicable not only to the cross-entropy loss function. Mean squared error loss, which is a widely used loss function in several semi-supervised models~\cite{chen2021semisupervised}, is also compatible as follows:

\begin{equation}
    \label{eq:vcloss}
    \mathcal{L}_{VC-MSE} = \sum_{i=0,i \notin PC}^{K+1}(\sigma(l^{i})-t^{i})^2,
\end{equation} where $\sigma$ is the sigmoid function, $t^i$ is the binary objective label. $t^i=1$ if the sample is of the category $i$. Otherwise, $t^i=0$. The objective label of VC is $t^K=1$ if the sample is confusing.

\subsection{Virtual Weight}
%

As mentioned before, the weight vectors in a linear classifier can be decoupled into two parts: the direction and the magnitude. To make the virtual weight $w^{v}$, the most intuitive solution is to directly use the direction of the teacher feature vector $f^t$ and scale it with a magnitude factor (the minimal norm of the pre-define classifier weights is adopted):
\begin{equation}
	w^v = \frac{\hat f}{\parallel \hat f \parallel_{2}} * norm,
\label{eq:vcvw}
\end{equation}which is used in our former ECCV publication~\cite{chen2022semisupervised}. In this paper, we further explore a new option.

\begin{figure}[t]
\centering
\includegraphics[width=0.5\columnwidth]{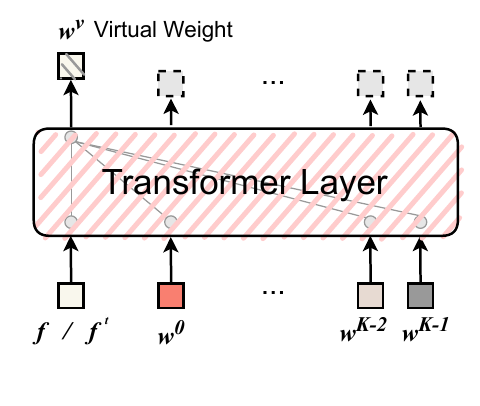}
\caption{The transformer layer for the virtual weight generation.}
\label{fig:transformer}

\end{figure}

As shown in \cref{fig:transformer}, a self-attention transformer layer is adopted to generate a learnable virtual weight. The input tokens consist of the weight vectors $w^{0 \sim K}$ in the student classifier and the feature vector $f$ or $f^t$. We use the first output of the transformer layer as the virtual weight $w^v$. This module is trained by the available ground truth and the pseudo labels with high confidence scores. By doing so, the weight generator is expected to be aware of the inherent relation between $w^{0 \sim K}$ and the feature vector to produce the final virtual weight.

\subsection{Potential Category Set}
\label{sec:pcset}
The potential category set consists of different predictions of one training sample. Any method that can give reasonably different predictions can be used to build the potential category set. For example, inspired by consistency regularisation, the predictions under different conditions reveal the potential categories of a confusing sample. The pseudo label inferring process can be represented by: $y'=g_{\theta}(x^u)$, where $g$ is a neural network parameterised by $\theta$, and $x^u$ is an unlabelled input image. By adopting different $g_{\theta}$, predictions under different conditions, which constitute the PC set, can be obtained. The size of the potential category set is used to determine whether a training sample is confusing to the model or not. If different predictions agree with each other, \ie the number of elements of the potential category set is one, the training sample is not considered a confusing sample. Otherwise, it is a confusing sample. In this paper, different policies are tailored and investigated for different tasks based on the specificity of each task.

\subsubsection{Semantic segmentation}
\label{sec:pcset-seg}
%

\noindent\textbf{Top-2 probability.} Most semi-supervised semantic segmentors usually adopt a high confidence score threshold, such as 0.95, to filter out confusing pixels. However, there are still many valuable low-confidence pixels. We use the categories of the top-2 probability to create the potential category set as the simplest policy for them in semantic segmentation.

\noindent\textbf{Teacher-student mutual verification.} Though the parameters of the teacher model are updated by the student's parameters, comparing the segmentation result $y^{'(s)}$ inferred by the student with $y^{'(t)}$ produced by the teacher finds many confusing pixels in the pseudo labels of semantic segmentation.

\subsubsection{Object detection}
\label{sec:pcset-det}
Unlike semantic segmentation, the location of pseudo labels plays a crucial role in the model training of object detection. Therefore, the confidence score-based methods, such as the top-2 probability policy, cannot be used in object detection since a pseudo bounding box with a low confidence score usually means its position is also unreliable. The prediction format for object detection consists of the coordinates and categories of bounding boxes. The principle of creating a PC set in object detection is comparing two individual prediction sets of one image. Two sets of predictions may have varying numbers of bounding boxes located at different positions. Consequently, we employ the Intersection over Union (IoU) metric to match bounding boxes between the two sets. When the IoU between two boxes exceeds a threshold (set at 0.5 in this paper), we compare their category predictions to determine if they are confusing samples. A bounding box in one set that doesn't find a match in the other set is considered a confusing sample. For example, if the category of a bounding box in one set is `bear', but finds no matching in the other set, the PC set is $\{bear, bg\}$.

We propose two easy-to-implement methods for object detection to get the two sets for the comparison:

\noindent\textbf{Temporal stability.} In object detection, the pseudo label of an image varies at different training iteration steps~\cite{yang2021interactive}. When pseudo-labels at different training steps are compared, those mismatched pseudo labels reveal the potential categories. We select the model of the current iteration step $g_{\theta^{cur}}$ and $g_{\theta^{last}}$ which is the checkpoint when the model viewed the current image the last time to produce two prediction sets $y^{'(1)}$ and $y^{'(2)}$ for the comparison. In the first training epoch, VC learning does not involve any data, as the model views all images for the first time.

\noindent\textbf{Cross-model verification.} Comparing the decisions of two conditionally independent models $g_{\theta^{(a)}}$ and $g_{\theta^{(b)}}$ for the same sample $x$ can also be used to discover the potential categories. Two models are initialised with different initial parameters. The orders of the training data for these two models are also different, ensuring that they do not collapse on each other \ie $\theta^{(a)} \neq \theta^{(b)}$. 

\vspace{1em}

\noindent From the aspect of the design of PC set creation, the PC set creation method for object detection can also be used in semantic segmentation. The reason for using different methods for different tasks is rooted in the variances in the implementation aspects of the two tasks. For example, the cross-model verification in semantic segmentation requires an extremely high GPU memory to train two deep segmentation models simultaneously.

In summary, if the size of the potential category set is not equal to 1, it means that this sample should be considered as a confusing sample. The VC learning will take over the training of this confusing sample. Regular loss, such as the cross-entropy loss, is used for unambiguous samples.

\begin{algorithm}[t]
\caption{The training process with VC learning.}\label{code-vc}
\definecolor{codeblue}{rgb}{0.25,0.5,0.5}

\algcomment{\fontsize{7.2pt}{0em}\selectfont \texttt{einsum}: sums the product based on the Einstein summation convention; \\ \texttt{masked\_softmax}: softmax with ignoring index argument; \\ \texttt{cat}: concatenation.
}

\lstset{
  backgroundcolor=\color{white},
  basicstyle=\fontsize{7.2pt}{7.2pt}\ttfamily\selectfont,
  columns=fullflexible,
  breaklines=true,
  captionpos=b,
  commentstyle=\fontsize{7.2pt}{7.2pt}\color{codeblue},
  keywordstyle=\fontsize{7.2pt}{7.2pt},
}
\begin{lstlisting}[language=Python]
"""

x:              Tensor(1*3*H*W), unlabelled images
model_t:        Teacher model
model_s:        Student model
loss_fn:        Classification loss function

"""

x_w = aug(x)
x_s = strongaug(x)
  
# Get logits of K classes and student features
logits, feats = model_s(x_s)
  
# Get pseudo labels and teacher features
logits_1, feats_1 = model_t(x_w)
logits_2, feats_2 = model_s(x_w)
y_1, y_2 = argmax(logits_1), argmax(logits_2)
  
if y_1 == y_2:
    return loss_fn(softmax(logits), y_1)
    
# Get logit of virtual category
logit_vc = einsum(feats, vw_gen(feats_1), "nc,nc->n")
logits = cat((logits, logit_vc), dim=1)
  
# Masked Softmax, ignore potential categories.
probs, mask = masked_softmax(logits, ign_idx=[y_1, y_2])
loss = loss_fn(
            probs, num_classes, 
            reduction="none") * mask
loss = loss.sum(dim=1).mean()
  
return loss
\end{lstlisting}

\end{algorithm}

\section{Experiments}\label{sec:exp}

\begin{table}[t]
\centering
\caption{Ablation study on different $t_{low}$. The values inside the parentheses indicate peak performance.}
\begin{tabular}{p{0.1\columnwidth}|p{0.22\columnwidth}|p{0.22\columnwidth}|p{0.22\columnwidth}}
	\toprule
	$t_{low}$ & 0.0 & 0.3 & 0.6 \\
	\midrule
	mIoU &  49.02 (49.02) & 49.25 (49.25) & 48.94 (49.29) \\
	\bottomrule
\end{tabular}
\label{tab:ablation-r1-q7}
\end{table}

In this section, experiments on two dense prediction tasks --- semantic segmentation and object detection --- are conducted to evaluate the proposed VC learning.

\subsection{Implementation Details}
The teacher-student pseudo labelling method serves as the baseline in this section. Given an unlabelled image $x^u$, the softmax output $P(x^u)$ is inferred by the teacher model. In the semantic segmentation task, $P(x^u)$ is a pixel-level probability matrix. While in object detection, it is an instance-level probability distribution. A predefined confidence threshold $t_{low}$ is used to filter out extremely noisy pseudo labels using $max(P(x^u)) < t_{low}$. The prediction with a very high confidence score $max(P(x^u)) > t$ will be fully trusted. We ablate this threshold and found that the final performance does not show significant differences (see \cref{tab:ablation-r1-q7}). But with a threshold of 0.6 (we adopt in this paper), the convergence speed is faster. The remaining pseudo labels and all available ground truth labels participate in the semi-supervised training.  The proposed method is implemented in the PyTorch framework~\cite{paszke2019pytorch}. The code can be found at the public repository\footnote{\hyperlink{https://github.com/GeoffreyChen777/VC}{https://github.com/GeoffreyChen777/VC}}.


The training process with VC learning is shown in \cref{code-vc}. Here we take a general classification model and the teacher-student mutual verification as an example for simplicity. If the pseudo labels produced by the teacher and the student agree with each other, VC learning returns the loss value without any additional process. Otherwise, the logit of the potential category will be ignored when the masked softmax function is performed. Finally, the target of the loss function is replaced by the virtual category (here it is K in \cref{code-vc}) to calculate the loss value.

\subsection{Semantic Segmentation}

\begin{table}[t]
\centering
\footnotesize
\caption{The hyperparameters and augmentation settings of VC learning in semi-supervised semantic segmentation.}
\begin{tabular}{m{0.37\columnwidth}|m{0.55\columnwidth}}
\toprule
\textbf{Hyper-parameter} &  \textbf{Value} \\
\midrule
$t$ & 0.95 \\
$t_{low}$ & 0.6 (Pascal VOC) / 0.8 (Cityscapes)\\
VC loss type & CE \\
EMA momentum & 0.9996 \\
\midrule
optimiser & SGD \\
learning rate $lr$ & 0.001 (Pascal VOC) / 0.002 (Cityscapes) \\
weight decay & 0.0001 \\
momentum & 0.9 \\
iteration num & 40K (Pascal VOC) / 80K (Cityscapes) \\
$\beta$ & 1 \\
\midrule
labelled data number & 1/64, 1/96, 1/128 (Pascal VOC) \\
& 1/64, 1/72, 1/80 (Cityscapes) \\
batch size (labelled) $bs^l$ & 4 \\
batch size (unlabelled) $bs^u$ & 4 \\
\midrule\midrule
\textbf{Augmentation} & \textbf{Parameters} \\
\midrule
\multicolumn{2}{l}{weak augmentations} \\
\midrule
Random Flip & $p=0.5$ \\
Random Crop & 512$\times$512 \\
Random Resize & Scales: [0.5, 0.75, 1, 1.5, 1.75, 2.0] \\
\midrule
\multicolumn{2}{l}{strong augmentations} \\
\midrule
weak aug. & same as above \\
Random ColorJitter &  \\
Random Grayscale & $p=0.2$ \\
Random Gaussian Blur & $p=0.5$ \\
Cutout & $p=0.5$ s=(0.02, 0.4), r=(0.3, 3.3) \\
\bottomrule
\end{tabular}
\label{tab:seg-hyperparams}
\end{table}

\subsubsection{Datasets and Evaluation Protocol}
For the semantic segmentation task, we perform experiments on two well-known datasets, Pascal VOC and Cityscape. PascalVOC 2012 consists of 20 classes and one background class. The size of the training images is 1464. Hariharan \etal\cite{hariharan2011semantic} augment PascalVOC 2012 with 9118 additional images. All the 10582 images are adopted in our experiments following the mainstream settings of the semi-supervised segmentation community. Cityscape contains 5k pixel-level labelled images of urban street scenes. The evaluation metric is the mean of Intersection over Union (mIoU).
\subsubsection{Settings}

The semantic segmentation model of all the following experiments is DeeplabV3+~\cite{chen2018encoderdecoder} with a ResNet50~\cite{he2016deep} backbone. We randomly divide the dataset with partitions 1/64, 1/96, and 1/128 for labelled/unlabelled data in Pascal VOC, respectively. 1/64, 1/72, and 1/80 are adopted for Cityscapes. The overall loss function is as follows:
\begin{equation}
	\mathcal{L} = \mathcal{L}_{l}(x^l;y)+\beta\mathcal{L}_{VC}(x^u;y').
\end{equation} Other details of the model set-up are shown in \cref{tab:seg-hyperparams}.

\begin{table*}[t]
\centering
\caption{The performance of semantic segmentation on Pascal VOC with different label ratios.}
\begin{tabular}{p{0.18\textwidth}|P{0.04\textwidth}P{0.04\textwidth}P{0.04\textwidth}P{0.04\textwidth}|P{0.04\textwidth}P{0.04\textwidth}P{0.04\textwidth}P{0.04\textwidth}|P{0.04\textwidth}P{0.04\textwidth}P{0.04\textwidth}P{0.04\textwidth}}

\toprule
Pascal VOC label ratio & \multicolumn{4}{c|}{1/64} & \multicolumn{4}{c|}{1/96} & \multicolumn{4}{c}{1/128} \\

Fold & 1 & 2 & 3 & mean & 1 & 2 & 3 & mean & 1 & 2 & 3 & mean \\
\midrule
CCT (CVPR20)~\cite{ouali2020semisupervised} & 39.12 & 42.82 & 50.65 & 44.20 & 37.81 & 38.79 & 33.19 & 36.60 & 23.89 & 23.89 & 28.22 & 25.33 \\
GCT (ECCV20)~\cite{ke2020guided} & 41.30 & 41.34 & 50.10 & 44.25 & 41.07 & 35.67 & 40.80 & 39.18 & 28.91 & 35.46 & 33.85 & 32.74\\
CPS (CVPR21)~\cite{chen2021semisupervised} & 45.34 & 44.15 & 47.83 & 45.78 & 40.06 & 33.78 & 37.99 & 37.28 & 29.71 & 32.54 & 31.22 & 31.16 \\
USRN (CVPR22)~\cite{guan2022unbiased} & 61.70 & 50.44 & 53.22 & 55.12 & 46.02 & 44.51 & 46.40 & 45.64 & 39.59 & 37.43 & N/A & 38.41 \\
ST++ (CVPR22)~\cite{yang2022st} & 62.11 & 59.39 & 62.43 & 61.31 & 55.57 & 54.46 & 55.35 & 55.13 & 45.54 & 46.99 & 43.42 & 45.32 \\
\midrule
Our Baseline & 62.91 & 60.45 & 57.61 & 60.32 & 56.01 & 57.84 & 55.86 & 56.57 & 46.08 & 53.40 & 52.83 & 50.77 \\
VC & \textbf{65.54} & \textbf{63.93} & \textbf{58.86} & \textbf{62.78} & \textbf{59.02} & \textbf{58.60} & \textbf{59.12} & \textbf{58.91} & \textbf{48.94} & \textbf{55.37} & \textbf{54.71} & \textbf{53.00} \\
\bottomrule

\end{tabular}
\label{tab:seg-voc}
\end{table*}

\begin{table*}[t]
\centering
\caption{The performance of semantic segmentation on Cityscape with different label ratios.}
\begin{tabular}{p{0.18\textwidth}|P{0.04\textwidth}P{0.04\textwidth}P{0.04\textwidth}P{0.04\textwidth}|P{0.04\textwidth}P{0.04\textwidth}P{0.04\textwidth}P{0.04\textwidth}|P{0.04\textwidth}P{0.04\textwidth}P{0.04\textwidth}P{0.04\textwidth}}

\toprule
Citysacapes label ratio & \multicolumn{4}{c|}{1/64} & \multicolumn{4}{c|}{1/72} & \multicolumn{4}{c}{1/80} \\

Fold & 1 & 2 & 3 & mean & 1 & 2 & 3 & mean & 1 & 2 & 3 & mean \\
\midrule
CCT (CVPR20)~\cite{ouali2020semisupervised} & 52.59 & 54.24 & \textbf{51.27} & 52.70 & 51.44 & 48.68 & 52.50 & 50.87 & \textbf{51.10} & 54.01 & 54.26 & 53.12 \\
GCT (ECCV20)~\cite{ke2020guided} & 52.14 & 51.52 & 49.20 & 50.95 & 47.43 & 46.62 & 52.23 & 48.76 & 47.46 & 51.06 & 51.31 & 49.94 \\
CPS (CVPR21)~\cite{chen2021semisupervised} & 46.80 & 48.05 & 48.02 & 47.62 & 42.10 &  46.69 & 48.16 & 45.65 & 46.36 & 40.63 & 44.89 & 43.96 \\
USRN (CVPR22)~\cite{guan2022unbiased} & N/A & 50.54 & N/A & 50.54 & N/A & N/A & N/A & N/A & N/A & N/A & 49.92 & 49.92 \\
ST++ (CVPR22)~\cite{yang2022st} & 51.67 & 56.56 & 49.11 & 52.45 & 51.20 & 47.02 & 52.77 & 50.33 & 44.02 & 49.64 & 46.04 & 46.57  \\
\midrule
Our Baseline & 51.72 & 56.63 & 49.84 & 52.73 & 53.38 & 48.96 & 51.20 & 51.18 & 48.35 & 59.59 & 53.81 & 53.92 \\
VC & \textbf{52.62} & \textbf{58.30} & 50.74 & \textbf{53.88} & \textbf{55.53} & \textbf{49.92} & \textbf{53.62} & \textbf{53.02} & 48.92 & \textbf{61.00} & \textbf{54.65} & \textbf{54.86} \\

\bottomrule

\end{tabular}
\label{tab:seg-city}
\end{table*}

\subsubsection{Baseline}

\begin{figure}[t]
\centering

\vspace{-1em}
\subfloat[]{
\includegraphics[width=0.49\columnwidth]{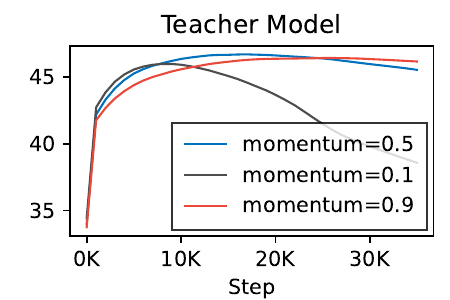}
\label{fig:baseline-seg-a}
}
\subfloat[]{
\includegraphics[width=0.49\columnwidth]{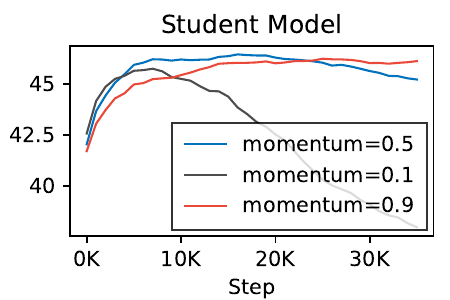}
\label{fig:baseline-seg-b}
}
\caption{The mIoU of the teacher-student baseline model with different BN momentums.}
\label{fig:baseline-seg}

\end{figure}

The teacher-student architecture (mean teacher~\cite{tarvainen2017mean}) is widely adopted in semi-supervised learning. However, it does not work satisfactorily in semantic segmentation. At the beginning of training, the model performance sees a remarkable improvement, but then the mIoU drops off a cliff. We find that the batch normalisation layer is the key component for a strong baseline model. In the limited-supervised setting, it can greatly affect the quality of pseudo labels and aggravate the confirmation bias issue.

A batch normalisation (BN) layer consists of two statistical parameters --- an estimated mean and an estimated standard deviation. However, different augmentations fit different BN parameters~\cite{yuan2021a}. The statistical information of the data used for pseudo labelling and model training is inconsistent. The pseudo labelling phase usually adopts a very basic augmentation to process the input data. In contrast, some strong augmentations are used for the data processing of the model training stage. Thus, we manage two groups of statistical parameters in the BN layers for each phase. 

Moreover, the estimated statistics are updated with a momentum argument by $x_{new}=(1-momentum)\times x_{old}+momentum\times x_{current}$\footnote{Here, the momentum is not the ema momentum to update the teacher parameters. It is the momentum in each BN layer for updating the estimated statistics.}. Most neural networks adopt a small momentum such as 0.1. As shown in~\cref{fig:baseline-seg}, we find that it leads to a severe training collapse when the numbers of labelled data are very limited. Increasing the momentum alleviates this issue. The possible reason is that a small momentum exacerbates a statistics bias due to the repeating sampling of the labelled data. Each training batch in a teacher-student framework is composed of labelled and unlabelled images. Due to the limited size of the labelled data subset, oversampling is quickly encountered. A small momentum makes the statistics of the labelled subset dominant in the estimated mean and standard deviation.

Thus, we adopt a relatively large momentum for all BN layers. Doing so makes the training stable, leading to an effective straightforward baseline model. The details can be found in the public source code repository.

\subsubsection{Performance}

We report the performance of our semi-supervised baseline and the model armed with the proposed VC learning in \cref{tab:seg-voc,tab:seg-city}. Since different selections of labelled images yield very different results in the extremely scarce-label setting, we randomly select three data folds to make the experiments more convincing. Other methods' results are reported in their papers or produced by their official codes. On the Cityscapes dataset, USRN cannot be optimised on some data folds due to the extremely unbalanced class distribution issue. Therefore, we use N/A in the table for those experimental results. To have a fair comparison, we use the final checkpoint to produce the results in \cref{tab:seg-voc,tab:seg-city} rather than the checkpoint with the highest mIoU. The strong teacher-student baseline model outperforms most of the state-of-the-art semi-supervised semantic segmentation algorithms. VC learning further boosts the baseline model to achieve a remarkable improvement. For example, on Pascal VOC fold 2 with a 1/128 label ratio, VC learning achieves a mIoU of 55.37, which is far higher than the recent ST++ (46.99). At most label ratios and data folds, VC learning surpasses others, indicating the superiority of VC learning.

In comparison to the VOC dataset, Cityscapes is a smaller (10k+ vs. 3k) and more challenging dataset. Adding some training data, such as increasing the label ratio from 1/80 to 1/64, does not significantly improve model performance. When we compare the three columns representing mean performance in \cref{tab:seg-city}, it's evident that most methods achieve similar mIoU scores across different label ratios. There is no clear, consistent trend of performance decline as the label ratio decreases, as observed with methods like CCT, GCT, and ours. However, ST++ significantly underperforms at extremely low label ratios, and one possible reason is overfitting. ST++ introduced a scheme to compare multiple-step predictions to filter out images with unstable predictions. Instead of utilising all available training data to the maximum extent, they discard a significant portion of it and retrain the model with the remaining data. This means that ST++ substantially reduces the number of training images, especially at very low label ratios. As a result, the model is prone to overfitting due to the training data being limited, leading to worse performance. In contrast, our VC learning makes an effort to leverage not only the confident training samples but also the confusing ones, resulting in the best overall performance.

We also evaluate VC learning with different backbones and report the segmentation results in \cref{tab:seg-backbone}. It is clear that VC learning constantly improves the baseline model with both backbones. On 1/128 label ratio, the baseline model with ResNet101 backbone performs worse than the one with ResNet50, which is possibly caused by overfitting.

\begin{figure}[t]
\centering

\subfloat[]{
\includegraphics[width=0.48\columnwidth]{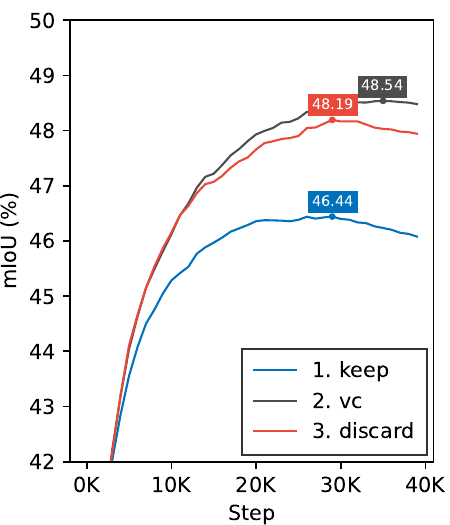}
\label{fig:ablation-seg-a}
}
\subfloat[]{
\includegraphics[width=0.48\columnwidth]{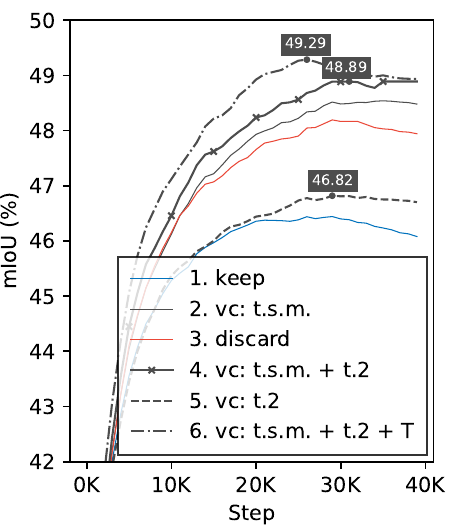}
\label{fig:ablation-seg-b}
}
\\
\vspace{-1em}
\subfloat[]{
\includegraphics[width=0.48\columnwidth]{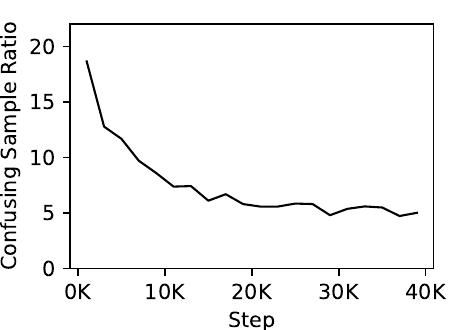}
\label{fig:ablation-seg-c}
}
\subfloat[]{
\includegraphics[width=0.48\columnwidth]{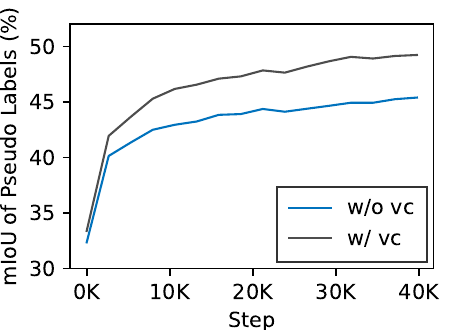}
\label{fig:ablation-seg-d}
}
\caption{ a) and b) Experiments of different strategies for dealing with confusing samples. t.s.m. means the teacher-student mutual verification policy for the potential category set creation. t.2 indicates the top-2 probability method. T means the virtual weight is produced by the weight generator formed by a transformer layer. c) Ratio of confusing samples. d) mIoU of pseudo labels w/ and w/o VC learning.}
\label{fig:ablation-seg}

\end{figure}

\subsubsection{Analysis and Ablation Study}

We ablate VC learning using the Pascal VOC 1/128 label ratio fold1 setting. By doing so, the value of the confusing sample and the effect of VC learning can be observed.

\begin{table}[t]
\centering
\caption{Segmentation results of our baseline model and VC learning with different backbones on Pascal VOC.}
\begin{tabular}{p{0.16\columnwidth}|p{0.18\columnwidth}|p{0.16\columnwidth}|p{0.16\columnwidth}}

\toprule
Backbone & Lable Ratio & Setting & mIoU \\
\midrule

\multirow{6}{*}{ResNet50} & \multirow{2}{*}{1/64} & baseline & 62.91  \\
&  & + VC & 65.54  \\ \cmidrule{2-4}
& \multirow{2}{*}{1/96} & baseline & 56.01  \\
&  & + VC & 59.02  \\ \cmidrule{2-4}
& \multirow{2}{*}{1/128} & baseline & 53.40 \\
&  & + VC & 55.37 \\

\midrule

\multirow{6}{*}{ResNet101} & \multirow{2}{*}{1/64} & baseline & 65.73  \\
&  & + VC & 66.11 \\ \cmidrule{2-4}
& \multirow{2}{*}{1/96} & baseline & 56.78  \\
&  & + VC & 60.03 \\ \cmidrule{2-4}
& \multirow{2}{*}{1/128} & baseline & 50.79 \\
&  & + VC & 55.23 \\

\bottomrule

\end{tabular}
\label{tab:seg-backbone}
\end{table}

\vspace{1em}
\noindent\textit{A. Ratio of confusing samples.}

\noindent The ratio of confusing samples is shown in \cref{fig:ablation-seg-c}. At the very beginning of training, the ratio of confusing samples is nearly 20\%. \cref{fig:ablation-seg-d} shows that the mIoU of the pseudo-labels in the early stage is unsurprisingly low. Thus, arbitrarily training the model with low-quality confusing samples makes it become a victim of confirmation bias. VC learning aims to alleviate this problem, thus resulting in better pseudo-label accuracy (w/ vc in \cref{fig:ablation-seg-d}).

\begin{table}[t]
\centering
\caption{Segmentation results of VC learning with CE loss and MSE loss on Pascal VOC.}
\begin{tabular}{p{0.18\columnwidth}|p{0.07\columnwidth}|P{0.11\columnwidth}P{0.11\columnwidth}P{0.11\columnwidth}P{0.11\columnwidth}}

\toprule
\multirow{2}{*}{Label Ratio} & \multirow{2}{*}{Loss} &  \multicolumn{4}{c}{Fold} \\
 &  &  1 & 2 & 3 & mean \\
\midrule
\multirow{2}{*}{1/64} & CE & 65.54 & 63.93 & 58.86 & 62.78 \\
 & MSE & 65.14 & 63.19 & 58.28 & 62.20 \\
 & NEG & 64.83 & 62.13 & 58.18 & 61.71 \\
\midrule
\multirow{2}{*}{1/96} & CE & 59.02 & 58.60 & 59.12 & 58.91 \\
 & MSE & 58.94 & 58.62 & 58.49 & 58.69 \\
 & NEG & 58.15 & 58.22 & 57.54 & 57.97 \\
\midrule
\multirow{2}{*}{1/128} & CE & 48.94 & 55.37 & 54.71 & 53.00 \\
 & MSE & 49.45 & 54.45 & 54.34 & 52.75 \\
 & NEG & 48.80 & 54.32 & 54.54 & 52.55 \\
\bottomrule

\end{tabular}
\label{tab:seg-mse-loss}
\end{table}

\vspace{1em}
\noindent\textit{B. Policies for confusing samples.}

\noindent In the previous section, we mentioned that neither discarding nor retaining is the optimal solution for confusing samples. In \cref{fig:ablation-seg-a}, we plot the mIoU curves of these two solutions and our VC learning. The keeping strategy (line 1, also to be regarded as the baseline) performs significantly worse than the other two solutions, which indicates the performance issues that arise due to  confirmation bias caused by wrong pseudo labels. By comparing line 3 with line 1, we can see that eliminating the influence of confirmation bias by discarding confusing samples can improve the model performance by far. However, it totally discards the potential contributions of confusing samples. Our VC learning (line 2) proactively utilises confusing samples, which further improves the mIoU to best the other runs. It supports our motivation for making use of confusing samples.

\vspace{1em}
\noindent\textit{C. Creating methods of potential category set.}

\noindent In \cref{fig:ablation-seg-a}, The potential category set creation method is the teacher-student mutual verification. It is only applied to the pseudo labels of high confidence pixels (\ie $> 0.95$). We also propose a method termed top-2 probability to deal with the low confidence pixels (\ie $< 0.95$) and plot the mIoU in \cref{fig:ablation-seg-b} as line 5. By comparing line 5 with line 1, it indicates that VC learning makes effective use of low-confidence data. We then combine the abovementioned two methods to gain an even better mIoU (line 4).

\vspace{1em}
\noindent\textit{D. Virtual weight.}

\noindent The previous ablation studies are conducted with the virtual weight proposed in our ECCV publication, \ie the normalised and scaled teacher feature vectors. This paper introduces an new version produced by a transformer layer. Line 6 in \cref{fig:ablation-seg-b} is the mIoU of VC learning with the virtual weight generated by the transformer. We can see that it exceeds all others, indicating that a better virtual weight is worth exploring in semantic segmentation. Moreover, we ablate the feature vectors from the teacher and the student as the input token of the transformer layer. A similar performance is observed: 48.94 ($f$) v.s. 48.76 ($f^t$).

To illustrate the individual contributions of direction and magnitude of the virtual weight, we employ cosine similarity as the loss function, replacing the cross-entropy (CE) and mean squared error (MSE) losses used in the basic version of VC learning. The results are presented in \cref{tab:ablation-r1-q1}.

\begin{table}[ht]

\caption{Cosine Sim. means using cosine similarity to replace the CE or MSE loss in VC learning.}
\begin{tabular}{p{0.2\columnwidth}|p{0.2\columnwidth}|p{0.2\columnwidth}|p{0.2\columnwidth}}
	\toprule
	Method & Baseline & Cosine Sim. & VC \\
	\midrule
	mIoU & 46.08 & 47.76 & \textbf{48.62} \\
	\bottomrule
\end{tabular}
\label{tab:ablation-r1-q1}

\end{table}

By maximizing the cosine similarity between the feature vector and the virtual weight while minimizing the cosine similarity between the feature vector and the classifier weight of negative categories, we disregard the magnitudes of both the virtual weight and the classifier weights. Cosine Sim. outperforms the baseline, highlighting the significance of the direction. In addition, it's important to note that VC demonstrates that magnitude is also crucial, as it achieves the best results.

\begin{figure}[t]
\centering

\includegraphics[width=0.9\columnwidth]{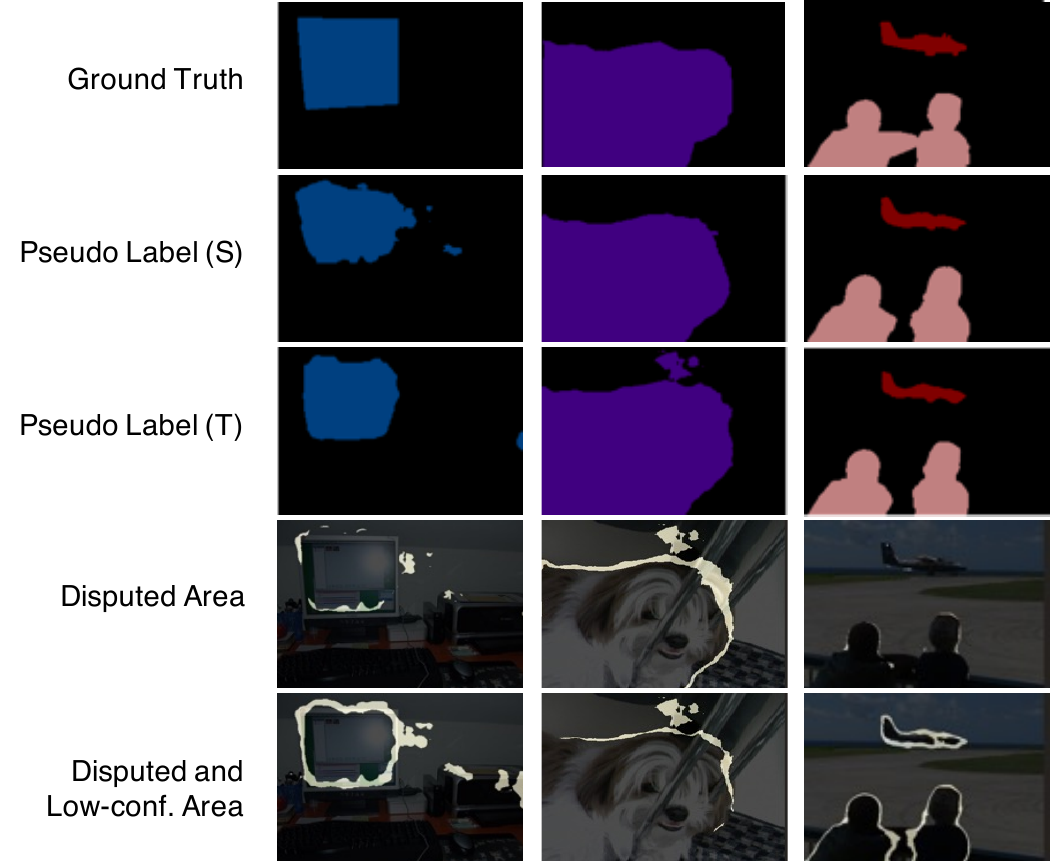}

\caption{Visualisation of potential category sets. The row of Pseudo Label (S) is inferred by the student model. The predictions of the teacher model are indicated by Pseudo Label (T).}
\label{fig:ablation-seg-vc-area}

\end{figure}

\begin{figure}[t]
\centering

\includegraphics[width=\columnwidth]{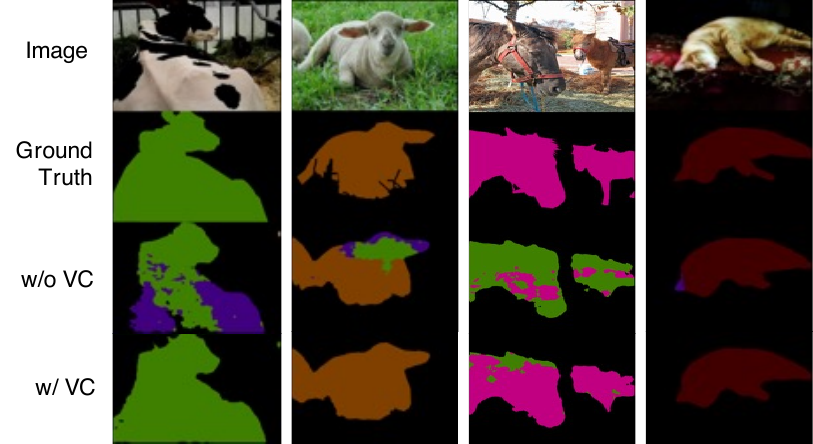}

\caption{Visualization results on the validation set of Pascal VOC.}
\label{fig:ablation-seg-results}

\end{figure}

\vspace{1em}
\noindent\textit{E. Visualisation of potential category set in semantic segmentation.}

\noindent To analyse what kind of areas VC learning will be mainly applied to, we visualise some demo pictures with their pseudo labels predicted by the student and teacher model in \cref{fig:ablation-seg-vc-area}. Disputed pixels are usually located on the boundaries of objects or semantically similar objects. For example, the model makes an indecisive decision about the monitor's boundary in column 1 of \cref{fig:ablation-seg-vc-area}. Some pixels belonging to the printer also introduce some false-positive predictions.

\vspace{1em}
\noindent\textit{F. Qualitative results.} 

\noindent The qualitative visual results of our baseline model and the model with VC learning are in~\cref{fig:ablation-seg-results}. The baseline model trained by limited labelled data produces many wrong predictions. On the contrary, the model with our VC learning processes these images very well.

\vspace{1em}
\noindent\textit{G. Loss forms.}

\noindent VC learning is compatible with not only the cross entropy (CE) loss function but also the mean squared error (MSE) loss function. The comparison results are reported in~\cref{tab:seg-mse-loss}. The VC learning with CE loss function outperforms the version with MSE loss. They all stand out against other methods, indicating the superiority and compatibility of our approach.

\vspace{1em}
\noindent\textit{H. Contribution of Virtual Category.}

\noindent In VC learning, we propose VC as the training label while omitting the categories in the potential category set. With the MSE loss, it is feasible to train the model with only the negative categories outside the potential category set. The performance is reported in \cref{tab:seg-mse-loss} of the label `NEG'. As indicated in \cref{tab:seg-mse-loss}, employing MSE with VC produces superior results compared to using only NEG. This highlights the substantial contribution of VC.

\subsection{Object Detection}

\subsubsection{Datasets and Evaluation Protocol}
To evaluate the proposed method for object detection, we assess it on two well-known object detection benchmark datasets -- MS COCO~\cite{lin2014microsoft} and Pascal VOC~\cite{everingham2015the}. Following the mainstream evaluation setting, we use the subset index provided by Unbiased Teacher~\cite{liu2021unbiased} to split the \textit{train set} across five different labelled ratios: 0.5\%, 1\%, 2\%, 5\% and 10\% (each ratio uses five random seeds used to obtain an averaged mAP). We also report the performance on Pascal VOC with \textit{VOC07-trainval} as the labelled subset and \textit{VOC12-trainval} as the unlabelled subset. Performance is evaluated on \textit{VOC07-test}. The evaluation metric for all the experiments reported in this subsection is mAP calculated by the COCO evaluation kit~\cite{coco_eval_kit}.

\begin{table}[t]
\centering
\footnotesize
\caption{The hyperparameters and augmentation settings of VC learning in semi-supervised object detection.}
\begin{tabular}{m{0.41\columnwidth}|m{0.5\columnwidth}}
\toprule
\textbf{Hyper-parameter} &  \textbf{Value} \\
\midrule
$t$ & 0.7 \\
VC norm &  const=3.5 \\
VC loss type & CE with focal loss term \\
PC set discovery & T.S. \\
EMA momentum & 0.9996 \\
\midrule
optimiser & SGD \\
learning rate $lr$ & 0.01 \\
weight decay & 0.02 \\
momentum & 0.9 \\
$\alpha$ & 4 \\
$\beta$ & 4 \\
iteration num & 180k \\
\midrule
labelled data number & 0.5-10\% (COCO) \\
& VOC07 as labelled data \\
batch size (labelled) $bs^l$ & 8 \\
batch size (unlabelled) $bs^u$ & 32 \\
\midrule\midrule
\textbf{Augmentation} & \textbf{Parameters} \\
\midrule
\multicolumn{2}{l}{weak augmentations} \\
\midrule
Random Flip & $p=0.5$ \\
Random Resize & Range: [400, 1200] \\
\midrule
\multicolumn{2}{l}{strong augmentations} \\
\midrule
weak aug. & same as above \\
Random ColorJitter &  \\
Random Grayscale & $p=0.2$ \\
Random Gaussian Blur & $p=0.5$ \\
Cutout & $p=0.7$ s=(0.05, 0.2), r=(0.3, 3.3) \\
& $p=0.5$ s=(0.02, 0.2), r=(0.1, 6.0) \\
& $p=0.3$ s=(0.02, 0.2), r=(0.05, 8.0) \\
\bottomrule
\end{tabular}
\label{tab:det-hyperparams}
\end{table}

\begin{table*}[t]
\centering
\caption{The performance on MS COCO with different label ratios. The results with $\dag$ are obtained from the available official code. Ours are the results with the same settings of Unbiased Teacher for the sake of fairness. Ours* is the results obtained by using some training settings in Soft Teacher. Different styles of underlines highlight the fair comparisons.}
\begin{tabular}{p{0.24\textwidth}|P{0.12\textwidth}|P{0.12\textwidth}|P{0.12\textwidth}|P{0.12\textwidth}|P{0.12\textwidth}}

\toprule
COCO label ratio &  0.5\% & 1\% & 2\% & 5\% & 10\% \\
\midrule
Supervised & 6.83 & 9.05 & 12.70 & 18.47 & 23.86\\
\midrule
CSD (NeurIPS19)~\cite{jeong2019consistencybased} & 7.41 & 10.51 & 13.93 & 18.63 & 22.46 \\
STAC (arXiv)~\cite{sohn2020a} & 9.78 & 13.97 & 18.25 & 24.38 & 28.64 \\
Instant Teaching (CVPR21)~\cite{zhou2021instantteaching} & - & 18.05 & 22.45 & 26.75 & 30.40 \\
Interactive (CVPR21)~\cite{yang2021interactive} & - & 18.88 & 22.43 & 26.37 & 30.53 \\
Humble Teacher (CVPR21)~\cite{tang2021humble} & - & 16.96 & 21.72 & 27.70 & 31.61 \\
Combating Noise (NeurIPS21)~\cite{wang2021combating} & - & 18.41 & 24.00 & 28.96 & 32.43 \\
\underline{Unbiased Teacher (ICLR21)}~\cite{liu2021unbiased} & 16.94 & 20.75 & 24.30 & 28.27 & 31.50 \\
\dashuline{Soft Teacher (ICCV21)}~\cite{xu2021endtoend} & 15.04\dag & 20.46 & 25.93\dag & 30.74 & 34.04 \\
MUM (CVPR22)~\cite{kim2022mum} & - & 21.27 & 26.84 & 31.90 & \textbf{35.92} \\
DTG (NeurIPS22)~\cite{li2022dtgssod} & 18.54 & 21.88 & 24.84 & 28.52 & 31.87 \\
\midrule
\underline{Ours} & 18.12 & 21.61 & 25.84 & 30.31 & 33.45 \\
\dashuline{Ours*} & \textbf{19.46} & \textbf{23.86} & \textbf{27.70} & \textbf{32.05} & 34.82 \\
\bottomrule
\end{tabular}
\label{tab:coco}
\end{table*}

\subsubsection{Settings}
Following the mainstream choice of the community, we adopted Faster-RCNN~\cite{ren2017faster} with FPN~\cite{lin2017feature} and ResNet-50~\cite{he2016deep} as the object detector. The training is conducted on 8 GPUs with batch size of 1/4 per GPU for labelled/unlabelled data. More details are introduced in \cref{tab:det-hyperparams}.

Object detection consists broadly of two subtasks: classification and localisation. Since the classification confidence score is not qualified to indicate the location quality of pseudo labels, some of the previous works~\cite{liu2021unbiased,sohn2020a} disabled the localisation loss of unlabelled data. We find that the method of creating the potential category set can also measure the quality of the location of pseudo labels. When we create the potential category set for a pseudo box $b$, we evaluate its location shift with the nearby box $\hat b$. We utilise the Intersection over Union (IoU) metric to match the bounding boxes. When the IoU between two boxes $b$ and $\hat b$ of two different predictions is higher than a predefined threshold (which we have set at 0.5 in this paper) and is the max one, $\hat b$ is considered the nearby box of $b$. We propose to decouple the horizontal and vertical boundary quality instead of using the IoU as a comprehensive metric to filter out the whole bounding box with low IoU value. The reason is that the IoU value can be affected by one biased boundary, even if the remaining boundaries are good. The horizontal quality flag $q_{hor}$ is calculated as:
\begin{equation}
    \label{eq:locflag}
    q_{hor}=\left\{
        \begin{array}{llll}
        1, & \frac{(x_1-\hat x_1)}{w} < t_{loc} & \& & \frac{(x_2-\hat x_2)}{w} < t_{loc}\\
        0, & otherwise & & \\
        \end{array} \right.,
\end{equation} where $x_1,x_2,\hat x_1,\hat x_2$ are the coordinates of the left and right boundary of the pseudo box $b$ and the nearby box $\hat b$, $w$ is the width of $b$, $t_{loc}$ is the threshold for high-quality boundaries. $q_{ver}$ is calculated in the same way. The decoupling allows high-quality boundaries to contribute to the localisation training. For example, the regression of the left and right boundary can be trained when the horizontal boundary quality is satisfied, even if the top and bottom boundaries are biased. The localisation loss consists of four Smooth-L1~\cite{girshick2015fast} loss terms:
\begin{equation}
    \label{eq:locloss}
    \mathcal{L}^{reg^*}=q_{hor}\mathcal{L}^x + q_{ver}\mathcal{L}^y + q_{hor}\mathcal{L}^w + q_{ver}\mathcal{L}^h.
\end{equation}

In summary, the overall loss function is as follows:

\begin{equation}
	\mathcal{L} = \mathcal{L}_{l}^{cls}(x^l;y) + \mathcal{L}_{l}^{reg}(x^l;y) +\beta\mathcal{L}_{VC}^{cls}(x^u;y')+\beta\mathcal{L}^{reg*}(x^u;y').
\end{equation} The classification loss for unlabelled data is replaced by the proposed VC learning loss term. The default hyperparameters and augmentation settings are as shown in \cref{tab:det-hyperparams}.

\subsubsection{Performance}

\noindent\textbf{MS COCO} We first evaluate our method on MS COCO with five label ratios. The results using 5 averaged random seeds are reported in~\cref{tab:coco}. The results with $\dag$ are obtained from the available official code. `Ours' are obtained by the model with exactly identical settings of UnbiasedTeacher, which is our baseline model. Given that scale jittering, as used in SoftTeacher, has been shown to have a substantial positive impact, we have incorporated this technique into our approach and report the corresponding results as `Ours*'. Furthermore, we adopt a relatively smaller batch size for labelled data, as recommended by SoftTeacher, to accelerate the training process in `Ours*'. The significant improvements can be summarised as follows:

\noindent 1) Compared with the supervised baseline, the mAP increases dramatically after training with the unlabelled data via our method.

\noindent 2) Our method outperforms other state-of-the-art semi-supervised detectors on all the label ratios by a significant margin. The mAP of our method at a small label ratio is close to or even exceeds the mAP of some methods using a large ratio.

\noindent\textbf{Pascal VOC} We also evaluate our method with VOC07 as the labelled subset and VOC12 and COCO* as the unlabelled subsets. We collect the images that contain objects in VOC predefined categories from MS COCO to build a subset \textit{COCO*}. The results are presented in~\cref{tab:voc}. Since the source codes of some methods are unavailable, the evaluation styles they used are unclear. Usually, the results based on the VOC-style AP are higher. Thus, we evaluate our method with both COCO-style mAP and VOC-style AP for the sake of fairness. Our method presents the best performance on these two unlabelled data subsets. Since VOC07 consists of more than 5K labelled images, and it is a relatively easy dataset,~\cref{tab:voc} indicates that our method can effectively further improve the performance, even if there is already sufficient labelled data.

\begin{table}[t]
\centering
\caption{Results of the experiment in VOC. $\mathcal{D}^l$ and $\mathcal{D}^u$ are the labelled and unlabelled subset choices. COCO* consists of the images from COCO that contain objects in the VOC categories. Numbers in () are obtained with the VOC-style AP.}
\begin{tabular}{m{0.25\columnwidth}|m{0.05\columnwidth}|m{0.2\columnwidth}<{\centering}|m{0.3\columnwidth}<{\centering}}

\toprule
\multirow{2}*{Method} & $\mathcal{D}^l$ & \multicolumn{2}{c}{VOC07}\\
 & $\mathcal{D}^u$ & VOC12 & VOC12 + COCO* \\
\midrule
\multicolumn{2}{l|}{STAC~\cite{sohn2020a}} & 44.64 & 46.01\\
\multicolumn{2}{l|}{Instant Teaching~\cite{zhou2021instantteaching}}& 50.00 & 50.80 \\
\multicolumn{2}{l|}{Interactive~\cite{yang2021interactive}} & 46.23 & 49.59\\
\multicolumn{2}{l|}{Humble Teacher~\cite{tang2021humble}} & 53.04 & 54.41 \\
\multicolumn{2}{l|}{Combating Noise~\cite{wang2021combating}} & 49.30 & 50.20  \\
\multicolumn{2}{l|}{Unbiased Teacher~\cite{liu2021unbiased}} & 48.69 & 50.34  \\
\multicolumn{2}{l|}{MUM~\cite{kim2022mum}} & 50.22 & \textbf{52.31}  \\
\midrule
\multicolumn{2}{l|}{Ours} & \textbf{50.40 (55.74)} & 51.44 (56.70) \\
\bottomrule
\end{tabular}
\label{tab:voc}
\end{table}

\subsubsection{Analysis and Ablation Study}
\label{sec:det-ablation}
In this subsection, we choose 1\% data of MS COCO as the labelled subset in object detection to analyse and validate our method in detail. All the experiments in this section are performed under the exactly same setting of the baseline model Unbiased Teacher except for the batch size. We adopt a smaller batch size to shorten the training time of each ablation study, therefore resulting in slightly decreased mAPs of all experiments compared to \cref{tab:coco}. The overall ablation study is reported in~\cref{tab:ablation-general}. The model with VC learning and Reg* Loss performs favourably against the baseline model.

\noindent\textit{A. Ratio of confusing samples.}

\begin{figure}[t]
\centering

\subfloat[]{
\includegraphics[width=0.48\columnwidth]{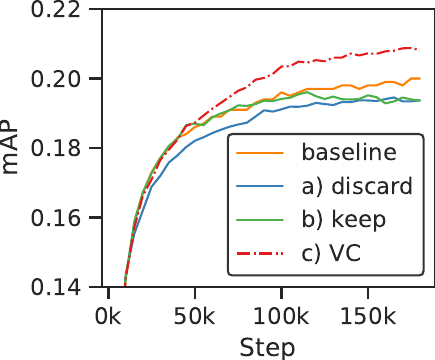}
\label{fig:ablation-det-1}
}
\subfloat[]{
\includegraphics[width=0.48\columnwidth]{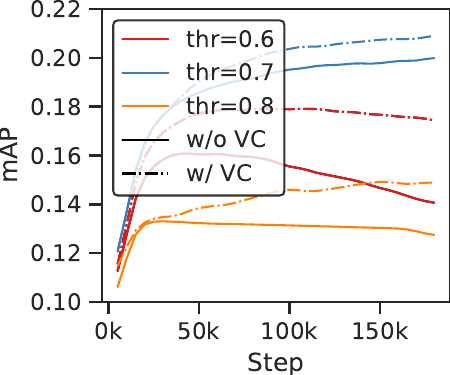}
\label{fig:ablation-det-2}
}
\\
\subfloat[]{
\includegraphics[width=0.48\columnwidth]{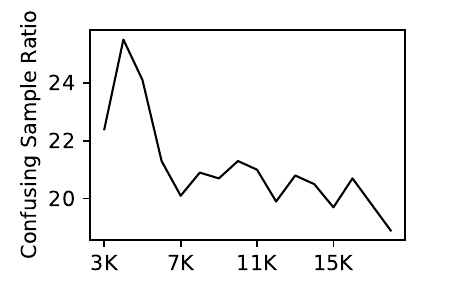}
\label{fig:ablation-det-3}
}
\subfloat[]{
\includegraphics[width=0.48\columnwidth]{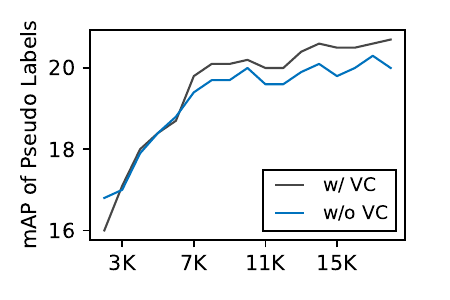}
\label{fig:ablation-det-4}
}
\caption{a) Experiments of different strategies for dealing with confusing samples. As the pseudo labels of the valuable confusing samples are highly unreliable, it is not optimal to either discard or keep them. Our VC learning(dot-dashed line) satisfies both demands, thereby resulting in a significant improvement. b) Experiments of different thresholds of the confidence score filtering w/ or w/o our VC learning. c) Ratio of confusing samples. d) mAP of pseudo labels w/ and w/o VC learning.}

\end{figure}

\noindent The ratio of confusing samples is shown in \cref{fig:ablation-det-3}. At the beginning of training, the ratio of confusing samples is increasing as there are many images the model encountered the first time. This ratio accounts for 20\% throughout the entire training process. This illustrates that there are numerous uncertain samples present when the number of labels is extremely limited. \cref{fig:ablation-det-4} shows the mAP of the pseudo-labels in the early stage, which reveals training the model with low-quality confusing samples yields a worse pseudo label quality, thereby hindering the model performance.

\vspace{1em}
\noindent\textit{B. Policies for confusing samples.}

\begin{table}[t]
\centering
\caption{Validation mAP with different strategies to deal with confusing samples.}
\begin{tabular}{p{0.15\columnwidth}|P{0.15\columnwidth}|P{0.15\columnwidth}|P{0.15\columnwidth}|P{0.15\columnwidth}}
	\toprule
	Strategy & baseline & a) discard & b) keep & c) VC \\
	\midrule
	mAP & 20.00  & 19.37 & 19.36 & \textbf{20.81} \\
	\bottomrule
\end{tabular}
\label{tab:ablation-vc}
\end{table}

\noindent Here, we adopt the temporal stability verification to create the potential category set for confusing samples. To analyse the effectiveness of the virtual category, we respectively report the mAP of the model under three policies: a) discarding all confusing samples (discard), b) retaining all potential labels for them (keep), and c) assigning our virtual category to replace the potential categories (VC). The baseline model is trained with vanilla pseudo labels (baseline) without the potential category discovery. As shown in~\cref{tab:ablation-vc}, both discarding and retaining policies decrease the mAP. By analysing the mAP during the entire training presented in~\cref{fig:ablation-det-1}, we noticed that rejecting confusing samples (blue line) results in a low mAP at the very beginning of the training. The reason is that using this policy discards some confusing samples with correct pseudo labels that the model needs. Then, as shown by the green line in~\cref{fig:ablation-det-1}, training with all potential categories gives a small performance boost at the early stage of training because more under-fitted samples are introduced to the model, but ends up with a low mAP. We believe this is due to the confirmation bias issue caused by incorrect pseudo labels that gradually hurts the performance. Our approach effectively resolves this conflict by providing a virtual category for the confusing sample.  The dot-dashed line in~\cref{fig:ablation-det-1} demonstrates that these confusing samples consistently benefit the model. The mAP sees a rise of 0.81 with our VC learning. The model with our VC learning exceeds the baseline early in the training and continues to lead until the end of the training. 

In addition, as can be seen from~\cref{fig:ablation-det-2}, we evaluate our VC learning with different thresholds (indicated by three colours) of the confidence score filtering adopted by our baseline model Unbiased Teacher. Confusing samples always exist, no matter whether the filtering mechanism is strict or not. The model with VC learning (dot-dashed lines) outperforms the baseline (solid lines) on three thresholds. Notably, the slump in the mAP disappears when $thr=0.6$, meaning that the confirmation bias has been effectively alleviated.

\begin{table}[t]
\centering
\caption{Ablation study on VC loss and modified localisation loss Reg* Loss. The method of creating the potential category set in VC learning is temporal stability verification.}
\begin{tabular}{P{0.2\columnwidth}|P{0.2\columnwidth}|P{0.2\columnwidth}}
	\toprule
		VC loss & Reg* Loss & mAP \\
	\midrule
		&  & 20.00 \\
	$\surd$ & & 20.81 \\
	$\surd$ & $\surd$ & \textbf{20.94} \\
	\bottomrule
\end{tabular}
\label{tab:ablation-general}
\end{table}

\vspace{1em}
\noindent\textit{C. Creating methods of potential category set}

\begin{table}[t]
\centering
\caption{Ablation study of different methods for creating the potential category. We also report the performance with only co-training techniques.}
\begin{tabular}{p{0.4\columnwidth}|P{0.3\columnwidth}}
\toprule
Method & mAP \\
\midrule
baseline & 20.00 \\
Temporal & 20.81 \\
Cross & \textbf{20.96} \\
co-training w/o VC & 20.53 \\
\bottomrule
\end{tabular}
\label{tab:ablation-pc}
\end{table}

\begin{table}[t]
\centering
\caption{Ablation study of different image augmentation for virtual weights generation.}
\begin{tabular}{p{0.25\columnwidth}|P{0.15\columnwidth}|P{0.15\columnwidth}|P{0.16\columnwidth}}
\toprule
Augmentation & none & flipping & strong aug. \\
\midrule
mAP & 20.80 & \textbf{20.81} & 20.70 \\
\bottomrule
\end{tabular}
\label{tab:ablation-aug}
\end{table}

\begin{table}[t]
\centering
\caption{Ablation study on the contributions of omitting the loss of categories in potential category set and VC.}
\begin{tabular}{p{0.1\columnwidth}|p{0.22\columnwidth}|p{0.22\columnwidth}|p{0.22\columnwidth}}
	\toprule
	Method & Omi. (discard) & Omi. (detach) & VC \\
	\midrule
	mIoU &  19.37 & 19.70 & \textbf{20.81} \\
	\bottomrule
\end{tabular}
\label{tab:ablation-r1-q4}
\end{table}

\begin{table}[t]
\centering

\caption{Ablation study of different hyperparameters. $t_{loc}$ is the threshold for location quality.}

\begin{tabular}{p{0.2\columnwidth}|P{0.2\columnwidth}|P{0.2\columnwidth}|P{0.2\columnwidth}}
\toprule
$t_{loc}$ & 0.03 & 0.05 & 0.1 \\
\midrule
mAP & 20.92 & 20.94 & 20.31\\
\bottomrule
\end{tabular}
\label{tab:ablation-hp-b}
\end{table}

\begin{figure}[t]
\centering

\includegraphics[width=\columnwidth]{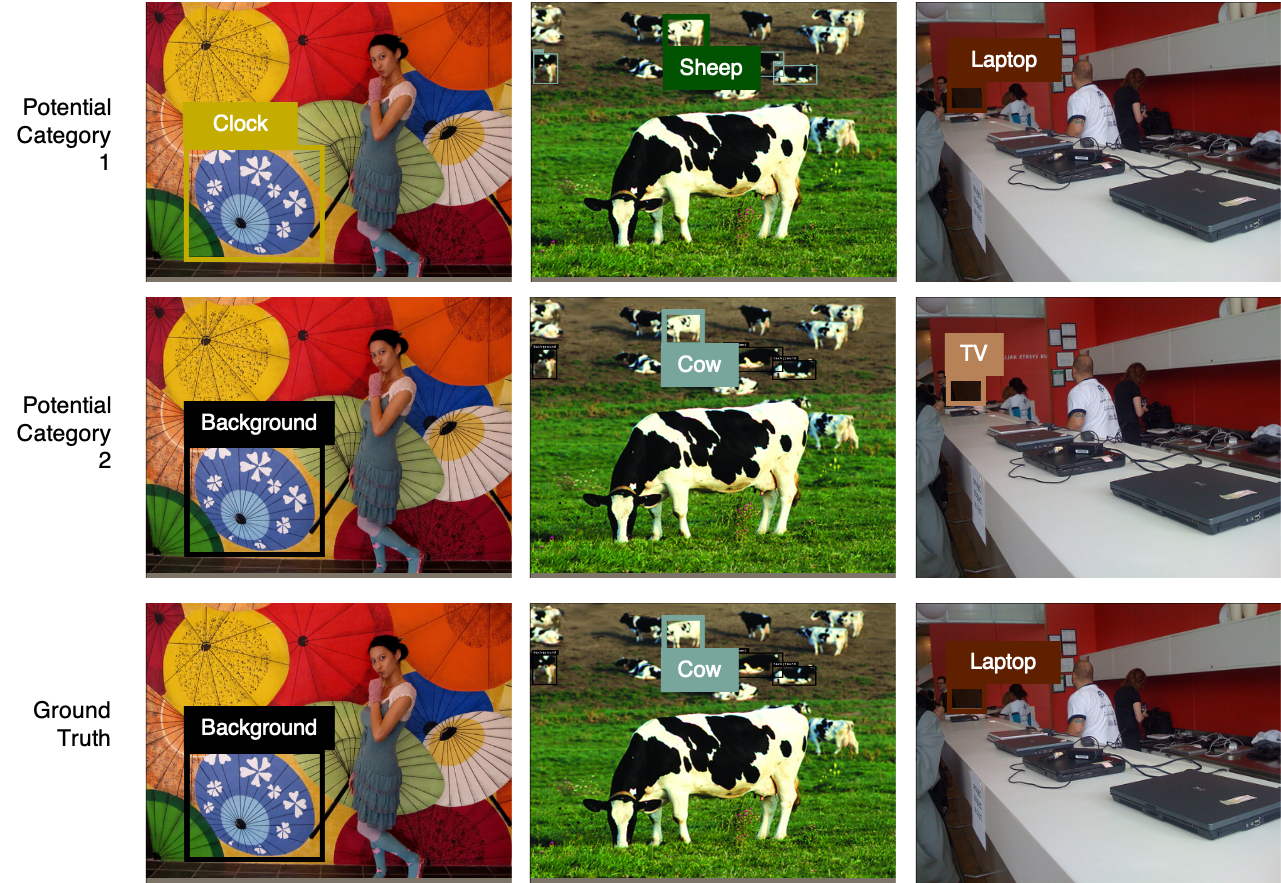}

\caption{Visualisation of potential category sets in object detection.}
\label{fig:ablation-det-pc-vis}

\end{figure}

\begin{table}[t]
\centering

\caption{The top-1 accuracy on miniImageNet (4k labels).}
\begin{tabular}{m{0.6\columnwidth}|m{0.3\columnwidth}}
\toprule
Method & Top-1 Acc. \\
\midrule
MeanTeacher~\cite{tarvainen2017mean} & 27.49 \\
Label Propagation~\cite{iscen2019label} & 29.71 \\
SimPLE~\cite{hu2021simple} & 49.39 \\
\midrule
Teacher-student Baseline & 43.10 \\
Ours & \textbf{51.49} \\
\bottomrule
\end{tabular}
\label{tab:acc-cls-res18}
\end{table}

\noindent\cref{sec:pcset-det} explored two methods to create the potential category set. We validate them and report the results in~\cref{tab:ablation-pc}. The cross-model verification achieves the best performance. The reason is that the cross-model verification is similar to the co-training technique which uses two independent models to provide pseudo labels for each other. It slightly alleviates the confirmation bias issue, thus resulting in additional improvement. As shown in~\cref{tab:ablation-pc}, the co-training can improve mAP by 0.53 individually. For the sake of fairness to other methods without co-training, we use the temporal stability verification in all other experiments, although the cross-model verification performs better.

\vspace{1em}
\noindent\textit{D. Virtual weight}

\noindent We choose the feature vector $f^t$ from the teacher model to produce the virtual weight. It is natural to validate different augmentations for the input image of the teacher model to generate various virtual weights. We explore three different settings: no augmentation, horizontal flipping, and strong augmentation. The results are reported in~\cref{tab:ablation-aug}. No performance gap can be observed between no augmentation and only horizontal flipping. Training with the virtual weight generated by the strong augmentation slightly degrades the mAP. The possible reason might be that the strong augmentation, especially the cutout, significantly perturbs the input image. Thus, in the feature space, the direction of the virtual weight is far away from the weight vector of the GT category.

In this paper, we introduce a transformer module to generate the virtual weight and its effect has been verified in the segmentation task. We also evaluate it in object detection on 1\% labelled COCO. The experiments show that it achieves comparable performance to the basic version (20.74 v.s. 20.81), \ie the feature vector from the teacher branch. Two main reasons are as follows: 1) unlike semantic segmentation that each pixel must have a label, the position of pseudo labels in object detection is highly uncertain, especially for the background class. Therefore, it is difficult to select a proper position and size of the bounding box to extract local features to train the transformer. The imbalance of foreground and background will also lead to the collapse of its training. 2) the quality of the virtual weight is not the bottleneck of VC learning in object detection. We evaluate the upper bound by using the ground truth to select the real weight as the virtual weight. It indicates that the improvement room is only 0.48 (21.29 vs 20.81). In summary, it is a good choice to adopt the most straightforward solution for object detection, \ie using $f^t$, which not only avoids the abovementioned problems but also achieves good results.

\vspace{1em}
\noindent\textit{E. Contribution of Virtual Category}

\noindent Similar to the ablation study in segmentation (Sec. 4.2.5 H). We conduct experiments to present the contribution of VC and the discarding operation. As shown in \cref{tab:ablation-r1-q4}, "Omi. (discard)" denotes that we remove all confusing training samples, while "Omi. (detach)" signifies that we omit the loss components associated with categories in the potential category set and disable the VC's contribution by detaching it from the PyTorch computational graphs. `Omi. (discard)' yields the worst results. This strategy discards all contributions from confusing samples, even though some correct pseudo-labels are undoubtedly conducive to optimization. `Omi. (detach)'  performs better than `Omi. (discard)'. This experiment confirms that the accuracy gain does not solely arise from ignoring the loss components of categories in the potential category set as it still falls short of the performance achieved by VC. `VC' delivers the best results. Our VC learning strives to leverage all confusing samples to the fullest extent possible, thereby achieving superior performance.

\vspace{1em}
\noindent\textit{F. Virtual Category learning in different stages}

\noindent We attempted to incorporate VC learning into the first stage of FasterRCNN, specifically the Region Proposal Network (RPN). Since the RPN is primarily engaged in a binary classification task, distinguishing between foreground and background regions, the confusion arises when determining whether a given region of interest corresponds to an object within the dataset or not. We observed that many of these confusing regions of interest pertain to out-of-dataset objects. Importantly, including such training samples for RPN does not negatively impact the final detection results, as the second stage of detection can effectively handle them. On the contrary, including these out-of-dataset objects during RPN training proves beneficial to some extent in identifying potential objects, especially when there are limited available ground truth labels. The experimental results are detailed in \cref{tab:ablation-r2-q6}.

\begin{table}[ht]
\centering
\caption{Ablation study on VC learning at different stages.}
\begin{tabular}{p{0.2\columnwidth}|p{0.2\columnwidth}|p{0.3\columnwidth}}
	\toprule
	Baseline & VC (stage 2) & VC (stage 1 and 2) \\
	\midrule
	20.00 & 20.81 & 20.72 \\

	\bottomrule
\end{tabular}
\label{tab:ablation-r2-q6}
\end{table}

\vspace{2em}

\vspace{1em}
\noindent\textit{G. Visualisation of potential category set.}

\noindent We visualise some demo pictures with their confusing pseudo labels in~\cref{fig:ablation-det-pc-vis}. Unlike semantic segmentation, labels for object detection can appear anywhere in an image. Therefore, the teacher model inevitably produces false positive pseudo labels as shown in the first column of~\cref{fig:ablation-det-pc-vis}. In addition, similar semantics also make the model fail to give accurate predictions.

\vspace{1em}
\noindent\textit{H. Hyperparameters of Reg* Loss}

\noindent The ablation study of the location quality threshold $t_{loc}$ in the $\mathcal{L}_{reg^*}$ are shown in~\cref{tab:ablation-hp-b}. A higher threshold will retain more unstable boundaries, leading to worse performance.

\subsection{Others}

In addition, we simply evaluate VC learning on miniImageNet~\cite{vinyals2016matching} for a non-dense task --- image classification. The top-1 accuracy of the proposed method is presented in \cref{tab:acc-cls-res18}. VC learning achieves 51.49\% on miniImageNet, surpassing existing works by a large margin.

\section{Discussion and Conclusion}

In this section, we discuss the limitations of VC learning first. The VC learning takes over the optimisation of the confusing samples. In object detection, by comparing the mAP gains of 10\% and other small label ratios, the improvement of our method is slightly lower but still rivals the first. This phenomenon is expected and reasonable. On the one hand, more labelled data means a better baseline detector. Thus, the room between the baseline and the fully-supervised upper bound is smaller. On the other hand, fewer unlabelled data and a better detector indicate that confusing samples are fewer. At the extreme, with 100\% labelled data, our VC learning will be applied to no sample, thereby resulting in no improvement. Notably, this scenario is not the topic of this paper. We focus on more practical situations, where very limited labelled data are available. The experiments with very limited labels demonstrated that the effectiveness of VC learning is remarkable. 

In conclusion, this paper proposed VC learning, which exploits the confusing under-fitted unlabelled data. We provide a virtual category label to a sample if its pseudo label is unreliable. It allows the model to be safely trained with confusing data for further improvement to achieve state-of-the-art performance. It can serve as a stepping stone to future work for the community of semi-supervised learning, especially with very limited labels.

\ifCLASSOPTIONcaptionsoff
  \newpage
\fi



%
\bibliographystyle{IEEEtran}
\bibliography{egbib.bib}
%

\begin{IEEEbiography}[{\includegraphics[width=1in,height=1.25in,clip,keepaspectratio]{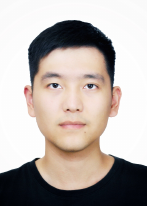}}]{Changrui Chen}
received the bachelor’s and master’s
degrees in computer science and technology from
the Ocean University of China in 2017 and
2020, respectively. He is currently pursuing the
Ph.D. degree with the WMG, University of Warwick, U.K.
His research interests are in computer vision and
semi-supervised learning.
\end{IEEEbiography}

\begin{IEEEbiography}[{\includegraphics[width=1in,height=1.25in,clip,keepaspectratio]{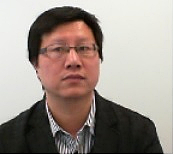}}]{Jungong Han}
is the Chair Professor in Computer Vision at the Department of Computer Science, University of Sheffield, U.K. He also holds an Honorary Professorship with the University of Warwick, U.K. His research interests include computer vision, artificial intelligence, and machine learning. He is a Fellow of the International Association of Pattern Recognition and serves as the Associate Editor for several prestigious journals, such as IEEE Transactions on Neural Networks and Learning Systems, IEEE Transactions on Circuits and Systems for Video Technology, and Pattern Recognition.
\end{IEEEbiography}


\begin{IEEEbiography}[{\includegraphics[width=1in,height=1.25in,clip,keepaspectratio]{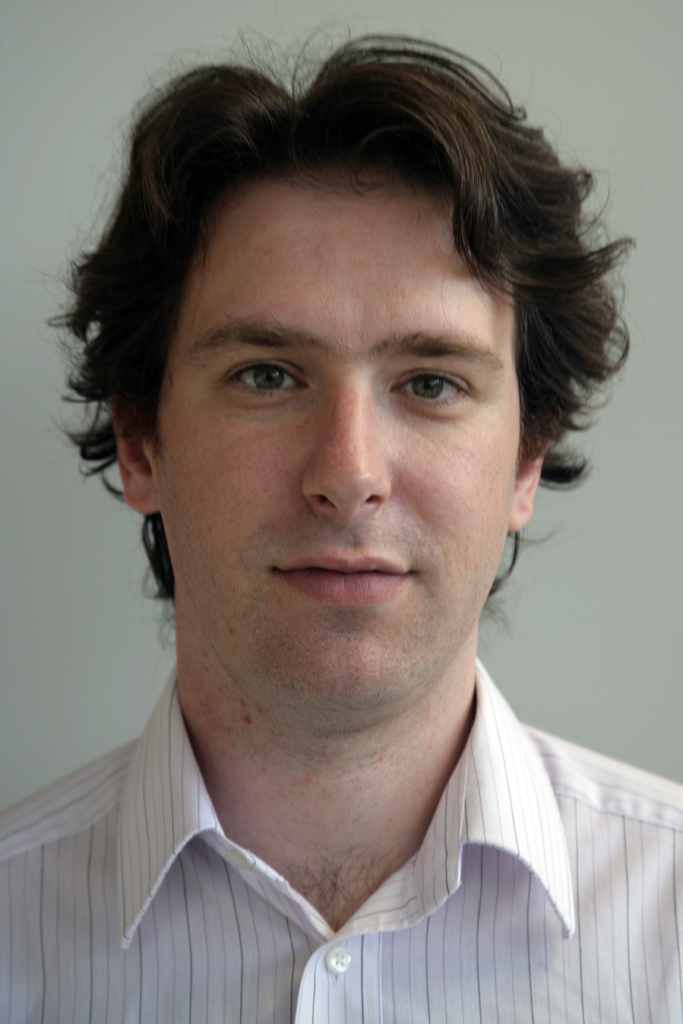}}]{Kurt Debattista}
Kurt Debattista received a B.Sc. in mathematics and computer science, an M.Sc. in psychology, an M.Sc. degree in computer science, and a Ph.D. from the University of Bristol. He is currently a Professor with WMG, at the University of Warwick. His research interests are high-fidelity rendering, HDR imaging, machine learning, and applied perception.
\end{IEEEbiography}





\end{document}